\documentclass[nohyperref]{article}

\makeatletter
\def\@xfootnote[#1]{%
  \protected@xdef\@thefnmark{#1}%
  \@footnotemark\@footnotetext}
\makeatother

\usepackage{microtype}
\usepackage{graphicx}
\usepackage{subfigure}
\usepackage{booktabs} % for professional tables

\usepackage{hyperref}
\usepackage{url}            % simple URL typesetting

\usepackage[accepted]{icml2022}

\usepackage{amsmath}
\usepackage{amssymb}
\usepackage{mathtools}
\usepackage{amsthm}

\usepackage{xcolor}         % colors
\usepackage{multirow}
\usepackage{amsmath,amsfonts,bm}

{}
{}
{}
{}

\def\eqref#1{equation~\ref{#1}}
\def\1{\bm{1}}

\def\vk{{\bm{k}}}

\def\vq{{\bm{q}}}

\def\vv{{\bm{v}}}

\def\vx{{\bm{x}}}
\def\vy{{\bm{y}}}

\def\mW{{\bm{W}}}

\DeclareMathAlphabet{\mathsfit}{\encodingdefault}{\sfdefault}{m}{sl}
\SetMathAlphabet{\mathsfit}{bold}{\encodingdefault}{\sfdefault}{bx}{n}

\usepackage{graphicx}
\usepackage[capitalize,noabbrev]{cleveref}

\newcommand{\repo}{\url{https://github.com/IDSIA/modern-srwm}}

\theoremstyle{plain}

\theoremstyle{definition}

\theoremstyle{remark}

\usepackage[textsize=tiny]{todonotes}

\icmltitlerunning{A Modern Self-Referential Weight Matrix That Learns to Modify Itself}

\begin{document}

\twocolumn[
\icmltitle{A Modern Self-Referential Weight Matrix That Learns to Modify Itself}

\begin{icmlauthorlist}
\icmlauthor{Kazuki Irie}{idsia}
\icmlauthor{Imanol Schlag}{idsia}
\icmlauthor{R\'obert Csord\'as}{idsia}
\icmlauthor{J\"urgen Schmidhuber}{idsia,kaust}
\end{icmlauthorlist}

\icmlaffiliation{idsia}{The Swiss AI Lab, IDSIA, USI \& SUPSI, Lugano, Switzerland}
\icmlaffiliation{kaust}{AI Initiative, King Abdullah University of Science and Technology (KAUST), Thuwal, Saudi Arabia}

\icmlcorrespondingauthor{}{\{kazuki, imanol, robert, juergen\}@idsia.ch}


% \begin{icmlauthorlist}
% \icmlauthor{Firstname1 Lastname1}{equal,yyy}
% \icmlauthor{Firstname2 Lastname2}{equal,yyy,comp}
% \icmlauthor{Firstname3 Lastname3}{comp}
% \end{icmlauthorlist}



\icmlkeywords{Machine Learning, ICML}

\vskip 0.3in
]



\printAffiliationsAndNotice{}  % leave blank if no need to mention equal contribution

\begin{abstract}
The weight matrix (WM) of a neural network (NN) is its program. The programs of many traditional NNs are learned through gradient descent in some error function, then remain fixed. The WM of a self-referential NN, however, can keep rapidly modifying all of itself during runtime. In principle, such NNs can meta-learn to learn, and meta-meta-learn to meta-learn to learn, and so on, in the sense of recursive self-improvement. While NN architectures potentially capable of implementing such behaviour have been proposed since the '90s, there have been few if any practical studies. Here we revisit such NNs, building upon recent successes of fast weight programmers and closely related linear Transformers. We propose a scalable self-referential WM (SRWM) that learns to use outer products and the delta update rule to modify itself. We evaluate our SRWM in supervised few-shot learning and in multi-task reinforcement learning with procedurally generated game environments. Our experiments demonstrate both practical applicability and competitive performance of the proposed SRWM.
Our code is public\footnote[$\dagger$]{\repo}.
\end{abstract}

\section{Introduction}
\label{sec:intro}

The program of a neural network (NN) is its weight matrix (WM) \citep{Schmidhuber:90diffenglish}.
With prediction tasks, for example,
starting from random values, an NN training procedure based on gradient descent might update the WM to minimize an error function that favors compression of given input-output observations \citep{solomonoff1964formal}. The WM becomes permanent once training ends, and its usefulness is evaluated with respect to its generalisation capability on yet unseen data.

Many environments, however, continue to evolve after training has halted (e.g., \citet{lazaridou2021pitfalls, lin2021clear}), and the test setting may deviate from training in ways that exceed the NN's generalisation capability. Then human intervention might be required to re-train or fine-tune the model.
Instead, more general and autonomous systems should learn to update their own programs in the light of new experience without such intervention.
Especially in multi-task learning and meta-learning (learning to learn; \citet{Schmidhuber:87long}), it may be useful to learn how to keep changing and fine-tuning  the model in a way that quickly adapts to various situations and new challenges \citep{hochreiter2001learning, FinnAL17}.

In principle, a WM could learn by itself a way of executing  rapid WM adaptations in task-dependent and context-dependent fashion through a generic mechanism for recursive self-modification.
Various self-modifying NNs have been proposed previously (see Sec.~\ref{sec:rel}).
Here we revisit the self-referential WM \citep{Schmidhuber:92selfref, Schmidhuber:93selfrefann, Schmidhuber:93selfreficann, Schmidhuber:93selfrefieee} from the '90s in the light of modern techniques for updating and generating weights.
In particular, we leverage mechanisms which are now well established in the context of Fast Weight Programmers (FWPs, \citet{Schmidhuber:91fastweights, schmidhuber1992learning, schmidhuber1993reducing}; reviewed in Sec.~\ref{sec:background}).
FWPs have recently seen advancements in terms of
performance and scalability, inspired by their formal equivalence \citep{schlag2021linear}
to linear variants \citep{katharopoulos2020transformers, choromanski2020rethinking, peng2021random} of the popular Transformer \citep{trafo}.

Here we derive a new type of self-referential WM (SRWM)
which naturally emerges as an extension to recent works on FWPs.
We evaluate the proposed SRWM in three settings.
We start by demonstrating that the proposed model
is effectively capable of generating useful self-modifications
by showing that the model achieves competitive performance on standard few-shot learning benchmarks.
Second, by extending the few-shot learning setting
to a sequential multi-task learning setting,
we test the SRWM's ability to sequentially adapt itself to changes of the task at runtime.
Finally, by using ProcGen \citep{CobbeHHS20}, we evaluate it
in a multi-task reinforcement learning (RL) setting with procedurally generated game environments.
Overall, we demonstrate both practical applicability and competitive performance of the proposed method.

\section{Background on Fast Weight Programmers}
\label{sec:background}
Here we briefly review the essential components
of fast weight programmers (FWPs; \citet{Schmidhuber:91fastweights, schmidhuber1992learning, schmidhuber1993reducing}) which our model is built upon (Sec.~\ref{sec:srm}).
In what follows, let $t$, $d_\text{in}$, $d_\text{out}$, and $d_\text{key}$ denote positive integers.
FWPs have a \textit{slow} NN which can rapidly modify weights of another \textit{fast} NN.
The concept has seen a recent revival, in particular in light of its direct formal connection \citep{schlag2021linear} to linear variants \citep{katharopoulos2020transformers, choromanski2020rethinking, peng2021random, irie2021going} of the popular Transformer \citep{trafo}
when the weight generation is based on outer products between keys and values generated by the slow NN  \citep{Schmidhuber:91fastweights}.
Recent work augmented the basic FWPs \citep{Schmidhuber:91fastweights, schmidhuber1992learning} with an improved elementary programming instruction or update rule invoked by the slow NN to reprogram the fast NN,
called \textit{delta update rule} (akin to the delta rule by \citet{widrow1960adaptive}). The resulting ``DeltaNet'' \citep{schlag2021linear}
is a general purpose auto-regressive NN with linear complexity w.r.t. input sequence length, which transforms
the input $\vx_t \in \mathbb{R}^{d_\text{in}}$ to the output $\vy_t \in \mathbb{R}^{d_\text{out}}$ as follows:
\begin{eqnarray}
\label{eq:proj}
\vk_t, \vv_t, \vq_t, \beta_t &=& \mW_\text{slow}\vx_t\\
\label{eq:retrieve}
\bar{\vv}_t &=& \mW_{t-1} \phi(\vk_t) \\
\label{eq:update}
\mW_t &=& \mW_{t-1} + \sigma(\beta_t)(\vv_t - \bar{\vv}_t) \otimes \phi(\vk_t) \\
\vy_t  &=& \mW_t \phi(\vq_t) \label{eq:fw_get}
\end{eqnarray}
where $\otimes$ denotes the outer product, $\sigma$ is a sigmoid function, and $\phi$ is
an element-wise activation function whose output elements are all positive and sum up to one (e.g.~softmax).
In Eq.~\ref{eq:proj}, the input $\vx_t$ is first projected to
key $\vk_t \in \mathbb{R}^{d_\text{key}}$, value $\vv_t \in \mathbb{R}^{d_\text{out}}$ , query $\vq_t \in \mathbb{R}^{d_\text{key}}$ vectors and a scalar $\beta_t \in \mathbb{R}$
using a trainable weight matrix $\mW_\text{slow} \in \mathbb{R}^{(d_\text{out} + 2 * d_\text{key} + 1) \times d_\text{in}}$.
The generated key vector $\vk_t$ and a learning rate $\beta_t$ (also generated by the slow NN) are used to update the \textit{fast weight matrix}
$\mW_{t-1}$ using the delta rule expressed in Eqs.~\ref{eq:retrieve}-\ref{eq:update}.
The fast weight matrix is typically initialized to zero i.e.~$\mW_0=0$.
The final output $\vy_t$ is obtained by querying the updated fast weight matrix $\mW_{t}$
using a generated query vector $\vq_t$ (Eq.~\ref{eq:fw_get}).
We note that the use of $\phi$ function for both writing to (Eq.~\ref{eq:update}) and reading from (Eqs.~\ref{eq:retrieve} and \ref{eq:fw_get})
the fast weights is crucial for stability when
the delta rule is used (we refer the readers to \citet{schlag2021linear} for detailed explanations).

In practice, we use the multi-head version \citep{trafo} of the computation above, that is, after the projection (Eq.~\ref{eq:proj}), the vectors $\vk_t$, $\vv_t$, $\vq_t$ are split into equally sized $H$ sub-vectors, and the operations in Eqs.~\ref{eq:retrieve}-\ref{eq:fw_get}
are conducted by $H$ computation heads independently.
Also, the layer above is typically used as a replacement for self-attention layers in a regular Transformer architecture \citep{trafo}
while preserving other components 
such as feedforward blocks,
layer-norm and residual connections across layers.

The slow NN or the \textit{programmer} (here a one-layer feedforward NN; Eq.~\ref{eq:proj}) with \textit{slow weights} $\mW_\text{slow}$ learns by gradient descent to
continuously modify or \textit{program} the fast NN (here also a one-layer feedforward NN; Eq.~\ref{eq:fw_get}) with fast weights $\mW_{t}$
as it continually receives a stream of inputs.
For further extensions of this concept to more complex slow and fast NN architectures such as recurrent NNs, we refer the readers to another recent study \citep{irie2021going}.

FWPs are of great interest from the perspective of context-sensitive information processing,
since the fast WM is completely context-dependent: while processing some sequence, a continually changing custom fast NN is built on the fly.

Here we leverage this mechanism
to design a new kind of FWP which programs itself.
It can be naturally derived from the operations
described above, resulting in a ``modern'' version of 
the self-referential weight matrix \citep{Schmidhuber:92selfref, Schmidhuber:93selfrefann, Schmidhuber:93selfreficann, Schmidhuber:93selfrefieee} of the '90s.

\begin{figure}[ht]
    \begin{center}
        \includegraphics[width=0.8\columnwidth]{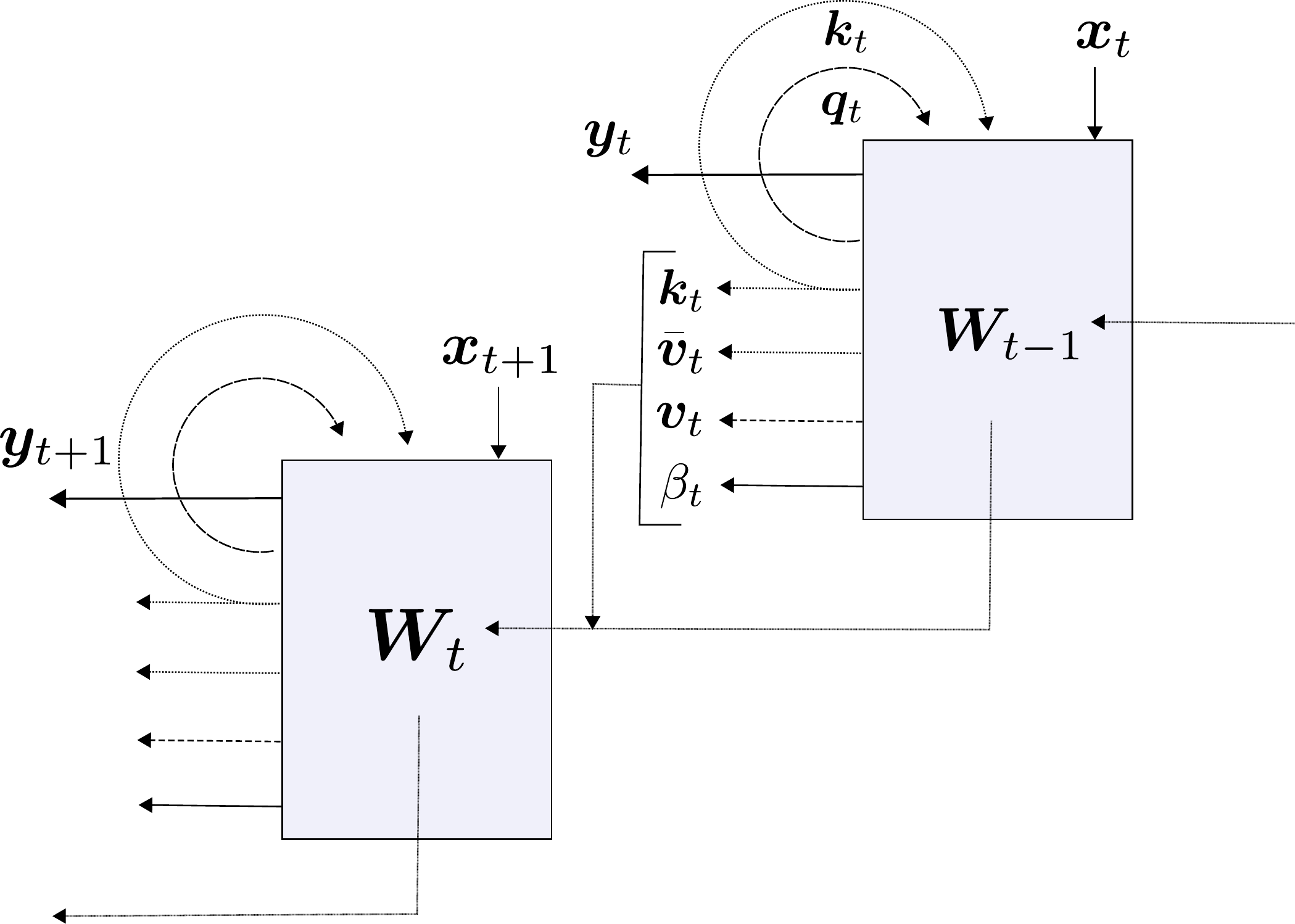}
        \caption{A ``modern'' self-referential weight matrix (SRWM).}
        \label{fig:srm}
    \end{center}
\end{figure}

\section{A Modern Self-Referential Weight Matrix}
\label{sec:srm}

Our ``modern'' self-referential weight matrix (SRWM) learns to train itself through self-invented key/value ``training'' patterns and learning rates, invoking sequences of elementary programming instructions based on outer products and the delta update rule, as in the recently proposed variants \citep{schlag2021linear} of FWPs (Sec.~\ref{sec:background}).

 Given an input $\vx_t \in \mathbb{R}^{d_\text{in}}$ at time $t$, our SRWM $\mW_{t-1} \in \mathbb{R}^{(d_\text{out} + 2 * d_\text{in} + 1) \times d_\text{in}}$ produces four variables $[\vy_t, \vq_t, \vk_t, \beta_t]$ where
 $\vy_t \in \mathbb{R}^{d_\text{out}}$ is the output of this layer at the current time step,
 $\vq_t \in \mathbb{R}^{d_\text{in}}$
 and $\vk_t \in \mathbb{R}^{d_\text{in}}$ are query and key vectors,
 and $\beta_t \in \mathbb{R}$ is the self-invented learning rate to be used
by the delta rule.
In analogy to the terminology introduced by the original SRWM papers \citep{Schmidhuber:92selfref, Schmidhuber:93selfrefann, Schmidhuber:93selfreficann, Schmidhuber:93selfrefieee},
$\vk_t \in \mathbb{R}^{d_\text{in}}$ is the \textit{modifier}-key vector, representing the key whose current value in the SRWM has to be modified, and
 $\vq_t \in \mathbb{R}^{d_\text{in}}$ is the \textit{analyser}-query
 which is again fed to the SRWM to retrieve a new ``value'' vector to be associated with the modifier-key.

The overall dynamics can be expressed as simply as follows:
\begin{align}
\label{eq:srm_start}
\vy_t, \vk_t, \vq_t, \beta_t &= \mW_{t-1} \phi(\vx_t) \\
\label{eq:srm_key}
\bar{\vv}_t &= \mW_{t-1} \phi(\vk_t) \\
\label{eq:srm_query}
\vv_t &= \mW_{t-1} \phi(\vq_t) \\
\label{eq:srm_end}
\mW_{t} &= \mW_{t-1} + \sigma(\beta_t)(\vv_t - \bar{\vv}_t) \otimes \phi(\vk_t)
\end{align}
where the value vectors have dimensions: $\vv_t, \bar{\vv}_t \in \mathbb{R}^{(d_\text{out} + 2 * d_\text{in} + 1)}$.
Figure \ref{fig:srm} illustrates the model.

Importantly, the initial values of the SRWM $\mW_0$ are the only
parameters in this layer which are trained by gradient descent.
In practice, we extend the output dimension of the matrix 
from ``3D+1'' $(d_\text{out} + 2 * d_\text{in} + 1)$ to ``3D+4'' $(d_\text{out} + 2 * d_\text{in} + 4)$
to generate four different, self-invented, time-varying learning rates
$\beta_t \in \mathbb{R}^4$ to be used in Eq.~\ref{eq:srm_end} for the four sub-matrices of $\mW_{t-1} = [\mW^y_{t-1}, \mW^q_{t-1}, \mW^k_{t-1}, \mW^\beta_{t-1}]$ used to produce
$\vy_t$, $\vq_t$, $\vk_t$, and $\beta_t$ in Eq.~\ref{eq:srm_start}.
For efficient computation, we also make use of multi-head computation
as is done in regular Transformers \citep{trafo, katharopoulos2020transformers}.
Please refer to Appendix \ref{app:model} for the full description.

The SRWM described above can potentially be used to replace any
regular WM.
Here we mainly focus on a model which can be obtained by replacing Eqs.~\ref{eq:proj}-\ref{eq:fw_get}
in the baseline DeltaNet
by the corresponding SRWM equations Eqs.~\ref{eq:srm_start}-\ref{eq:srm_end}.
In Appendix \ref{app:proc_mem}, we further show preliminary results for
another type of model incorporating an SRWM, which is also based on the DeltaNet but where we replace its slow weight matrix (Eq.~\ref{eq:proj}) by an SRWM (Eqs.~\ref{eq:srm_start}-\ref{eq:srm_end}).

\begin{figure}[t]
%    \vskip 0.2in
    \begin{center}
        \includegraphics[width=.85\columnwidth]{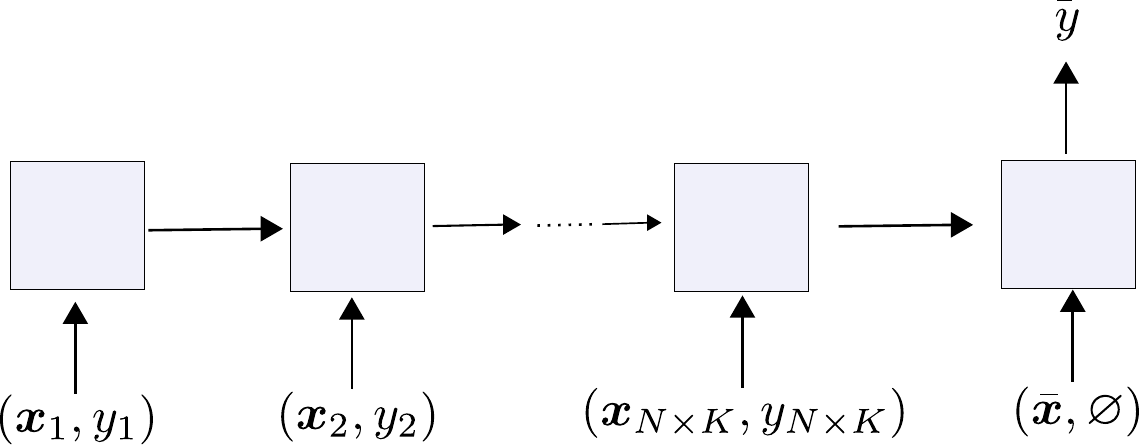}
        \caption{Synchronous-label setting for $N$-way $K$-shot learning (like in \citet{mishra2018a}). The correct label is fed together with the corresponding input for the first $N \times K$ tokens. Label prediction is only done for the ($NK$+1)-th input which is presented without the label.}
        \label{fig:sync}
    \end{center}
\end{figure}

\begin{table*}[t]
\caption{Single task, 5-way, few-shot classification test accuracies (\%) on Omniglot and Mini-ImageNet.
The bottom section shows the generic sequence models, while the top part shows other related approaches.
The numbers marked by * are taken from the corresponding papers. 
Following the standard convention \citep{RaviL17}, we report 95\% confidence interval computed over five sets of test episodes.
For further details, we refer to Appendix \ref{app:fs_exp}.
}
\label{tab:few_shot}
\vskip 0.15in
\begin{center}
%\begin{small}
\begin{tabular}{lclcc}
\toprule
       &   Omniglot  & \multicolumn{3}{c}{Mini-ImageNet} \\ \cmidrule(r){2-2} \cmidrule(r){3-5} 
        & 1-shot & Backend & 1-shot  & 5-shot  \\ \midrule
MAML* \citep{FinnAL17} &   98.7 $\pm$ 0.4 & Conv-4-32 & 48.7 $\pm$ 1.8 &  63.1 $\pm$ 0.9 \\
fwCNN-Hebb*  \citep{munkhdalai2018metalearning}  & \textbf{99.4} $\pm$ 0.1 &  Conv-5-64 & 50.2 $\pm$ 0.4 & 64.8 $\pm$ 0.5 \\
HyperTransformer* \citep{zhmoginov2022hypertransformer} &  \multicolumn{1}{l}{96.2} & Conv-4-32 & \multicolumn{1}{l}{\textbf{53.8}} & \multicolumn{1}{l}{\textbf{67.1}}\\
\midrule \midrule
SNAIL* \citep{mishra2018a} & \textbf{99.1} $\pm$ 0.2 & Conv-4-32   &  \multicolumn{1}{l}{45.1} & \multicolumn{1}{l}{55.2}  \\   % 99.07 $\pm$
Delta Net  & 97.2 $\pm$ 0.0 & Conv-4-32 & \textbf{47.0} $\pm$ 0.1 & \textbf{62.7} $\pm$ 0.1 \\
SRWM &  97.4 $\pm$ 0.0 & Conv-4-32 & \textbf{47.0} $\pm$ 0.2 & 61.4 $\pm$ 0.1 \\
\bottomrule
\end{tabular}
\end{center}
\vskip -0.1in
\end{table*}

\section{Experiments}
Can the self-referential dynamics described in Eqs.~\ref{eq:srm_start}-\ref{eq:srm_end}  generate useful self-modifications?
The overall goal of our experiments  is to evaluate
the proposed SRWM on various types of tasks
which require ``good" self-modifications.
We conduct experiments on standard supervised few-shot learning tasks (Sec.~\ref{sec:fsl}) and multi-task reinforcement learning (RL) in game environments (Sec.~\ref{sec:rl_exp}).
In addition, we show how our SRWM can be trained to efficiently adapt itself in a sequential multi-task few-shot learning setting  (Sec.~\ref{sec:seq_multi}).
Multi-task settings are particularly
relevant for evaluating SRWMs,
since unlike the baseline DeltaNet
(which uses the same fixed slow weights to generate fast weights for all tasks),
SRWMs can potentially learn to even adapt the way it adapts itself to each task as it receives task-specific inputs.

\begin{figure}[t]
    \begin{center}
        \includegraphics[width=.72\columnwidth]{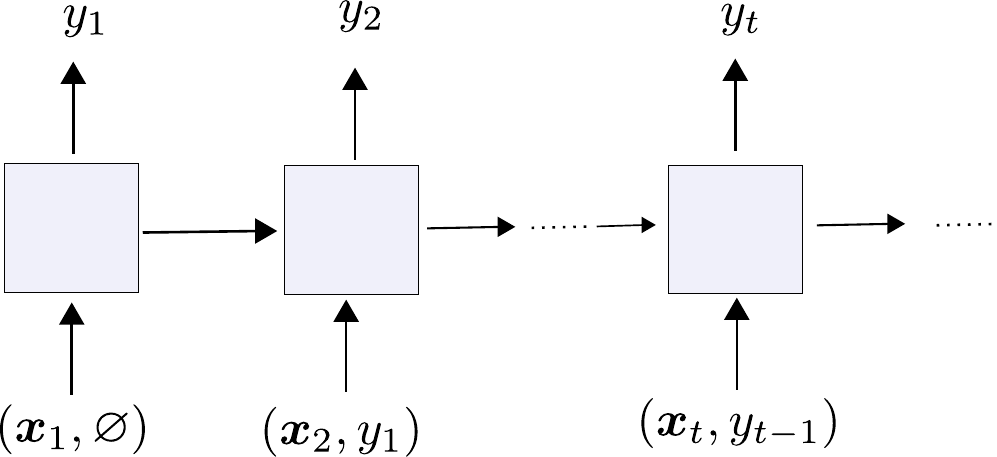}
        \caption{Delayed-label setting (like in \citet{hochreiter2001learning, SantoroBBWL16}). The correct label is fed one step after the corresponding input. Predictions take place at each step. }
        \label{fig:delayed}
    \end{center}
\end{figure}

\subsection{Standard Few-Shot Learning}
\label{sec:fsl}
We start with evaluating the proposed
SRWM's capability to generate useful self-modifications
on standard supervised few-shot image classification tasks.
We conduct experiments on the classic Omniglot \citep{lake2015human} and
Mini-ImageNet \citep{VinyalsBLKW16, RaviL17} datasets.
For further details on the datasets, we refer to the respective references and Appendix \ref{app:fs_exp} where we also provide
extra experimental results on the Fewshot-CIFAR100 \citep{OreshkinLL18} dataset.

Let $N$, $K$ and $C$ denote positive integers.
The task of few-shot image classification, or $N$-way $K$-shot image classification based on a dataset
containing $C$ classes, is structured through so-called \textit{episodes}.
In each episode, $N$ distinct classes are randomly drawn from $C$.
The resulting $N$ classes are re-labelled such that each class is assigned to one out of $N$ distinct random label indices.
For each of these $N$ classes, $K$ examples are randomly sampled.
The set of resulting $N \times K$ labelled images is called the \textit{support set}.
The goal of the task is to predict the label of another image (a \textit{query image} that is not in the support set) which is sampled from one of the $N$ classes,
based on the information available in the support set.

While there are several ways of approaching this problem, we are evaluating our SRWM
in the \textit{sequential learning} approach \citep{SantoroBBWL16, hochreiter2001learning}.
That is, the image/label pairs of the support set are randomly ordered
to form a sequence which is read by a sequence-processing NN (e.g., a recurrent NN).
By encoding the support set information into its internal state, the corresponding NN predicts the label of the query image.
In the case of our SRWM, the model generates updates of its own weights
as it reads the sequence of support set items.
The generated weights are used to compute the final prediction for the query image.
To fully specify this approach, we also need to explain how input image/label pairs are fed to the model.
Here we follow the approach used by \citet{mishra2018a} which we refer to as the \textit{synchronous-label setting} illustrated in Figure \ref{fig:sync}.
This strategy is specifically designed for $N$-way $K$-shot learning,
which consists in feeding the input and its label to the model at the same time for $N \times K$ items in the support set.
The model only predicts the label of the $(N \times K+1)$-th input which is the query image presented without label.
An alternative approach \citep{SantoroBBWL16} which we call the \textit{delayed label setting} (Figure \ref{fig:delayed})
is used later in Sec.~\ref{sec:seq_multi}.

\citet{mishra2018a}'s SNAIL model
serves as a baseline in our experiments
as it is a generic Transformer-like sequence processing model where the regular feedforward block is replaced by 1D convolution \citep{waibel1989phoneme}.
Other related approaches include the fwCNN-Hebb model by \citet{munkhdalai2018metalearning} and the recently proposed HyperTransformer \citep{zhmoginov2022hypertransformer}, as well as the standard MAML \citep{FinnAL17, FinnL18}.
Both fwCNN-Hebb and HyperTransformer generate fast weights of an NN based on the support set information.
The fwCNN-Hebb is an outer product based fast weight programmer (similar to ours but using the purely additive update rule like in the standard Linear Transformers) in which the fast net is a linear layer before the final softmax layer.
A crucial difference to our generic FWPs is that in the fwCNN-Hebb, key vectors are generated from the images while the values are generated from the labels, which is a specific design choice reflecting the task of few-shot learning.
In the HyperTransformers, the fast net is a convolutional NN whose weights are parameterised by the Transformer encoder architecture taking images and their labels as inputs.
We note that these two methods are specifically designed for few-shot learning,
unlike our models and SNAIL which are generic sequence processing NNs directly applicable beyond few-shot learning e.g., to RL as in Sec.~\ref{sec:rl_exp}.
MAML optimises the initial weights for future fine-tuning via gradient descent.
In the SRWM, the initial weights also learn and encode its own self-modification algorithm.
All models presented here use the same vision backend: the 4-layer convolutional net \citep{VinyalsBLKW16} with 32 channels for Mini-ImageNet (Conv-4-32), and 64 channels for Omniglot,
except the fwCNN-Hebb which uses 5 layers and 64 channels (Conv-5-64) in both cases.
Additional details of the model architecture are explained in Appendix \ref{app:fs_exp}.

The results are summarised in Table \ref{tab:few_shot}.
Overall, the proposed SRWM performs
well among the generic approaches.
Comparing the SRWM
to the SNAIL baseline,
the SRWM achieves very competitive performance on Mini-ImageNet\footnote{On Omniglot, we did not manage to reach $>$ 99.0\% performance of SNAIL.
We might be missing some technical details/tricks but
the results are nonetheless respectable.}.
DeltaNet and SRWM tend to have similar performance.
This is a satisfactory result, as it shows that a single self-modifying WM (instead of separate slow and fast nets) remains competitive in this single task scenario.

While we find the HyperTransformer to outperform all other models considered here,
our performance is respectable compared to that of MAML without requiring bi-level optimisation, and fwCNN-Hebb without inductive bias on the key/value generation from image/label, respectively.
Although the SRWM is a very generic approach, its overall performance is thus competitive,
demonstrating the effectiveness of the proposed self-referential dynamics (the main goal of this experiment).

\begin{table*}[h]
\caption{Total and instance-level accuracies (\%) for \textbf{sequential multi-task few-shot learning}
experiments (Sec.~\ref{sec:seq_multi}).
Regarding the instance-level accuracy, column $k \in \{1, 2, 3, 5, 10\}$ shows the percentage of correctly predicted $k$-th instances from each class.
This is for the test time scenario where the model is first tasked to learn to predict Omniglot and then Mini-ImageNet.
}
\label{tab:seq_adapt}
\begin{center}
\begin{tabular}{rr|rrrrr|r}
\toprule
Task & Model & 1 & 2 & 3 & 5 & 10 & Total \\ \midrule
\multirow{2}{*}{Omniglot}  & DeltaNet & 39.1 & 91.2 & 93.9 & 95.8 & 96.8 & 92.2 \\
         & SRWM & 40.6 & 92.1 & 94.8 & 96.3 & 96.7 & \textbf{92.3}  \\ \midrule

\multirow{2}{*}{Mini-ImageNet}  & DeltaNet & 20.8 & 43.5 & 48.2 & 51.3 & 54.8 & 50.4 \\
         & SRWM & 20.9 & 46.0 & 50.3 & 54.2 & 58.0 & \textbf{53.3} \\  
\bottomrule
\end{tabular}
\end{center}
\end{table*}

\subsection{Sequential Multi-Task Adaptation}
\label{sec:seq_multi}
The basic few-shot learning experiments presented in the previous section demonstrate that the very generic SRWM can effectively generate useful weight updates, achieving competitive performance on standard benchmarks.
Now we are interested in testing its self-referential dynamics on a task which requires adaptation to environmental changes at runtime.
We introduce two modifications to the few-shot learning
setting above.
First, instead of specifically training the model for $N$-way $K$-shot classification using the synchronous-label setting (Figure \ref{fig:sync}), we train our model
in the delayed-label setting \citep{hochreiter2001learning}
as illustrated in Figure \ref{fig:delayed}.
Here the model makes prediction at each time step
by receiving an input image to be classified and the correct label
of the previous input (the label feeds are thus shifted/delayed by one time step).
This setting is convenient for evaluating the model on a continuous stream of predictions/solutions\footnote{The synchronous-label setting could also be modified to support such usage, but that would require to feed the same input image twice, once without the correct label (for prediction) and once
with it (for learning). While such a setting may be of interest as a way of clearly separating the learning mode from the prediction mode, it is out of the scope of this work.}.
Second, the sequence of images to be predicted
is constructed by concatenating two image sequences  sampled from two different datasets: Omniglot and Mini-ImageNet.
The model first receives a stream of images
from one of the datasets; at some point, the dataset suddenly changes, to simulate a change of environment.
The model has to learn to adapt itself to this shift without human intervention, in a continual
execution of its program.

Note that our goal is to construct a task which requires adaptation to sudden changes during the model's runtime, which is different from \textit{continual few-shot learning} (e.g. \citet{yapRB21}) whose goal is to successively meta-train on multiple few-shot learning tasks.

We conduct experiments in
a 5-way classification setting with concatenation of Omniglot and Mini-ImageNet segments containing up to 15 examples per class in each segment.
The concatenation order is alternated for each batch,
and training segment lengths are randomised by trimming.
Regardless of model type, we find that training models in the delayed-label
setting is more difficult than in the synchronous-label setting.
We observe that in many configurations, the model gets stuck at a sub-optimal behaviour where it learns to improve its class-averaged zero-shot accuracy (apparently by learning to output one of the unused labels for a new class appearing in the sequence for the first time), but fails to learn to properly learn from the feedback at each step.
The most crucial hyper-parameter we identified was the large enough batch size.
Additional details on practicalities are shared in Appendix \ref{app:train_delayed}. 

\begin{figure}[h]
    \begin{center}
        \includegraphics[width=0.95\columnwidth]{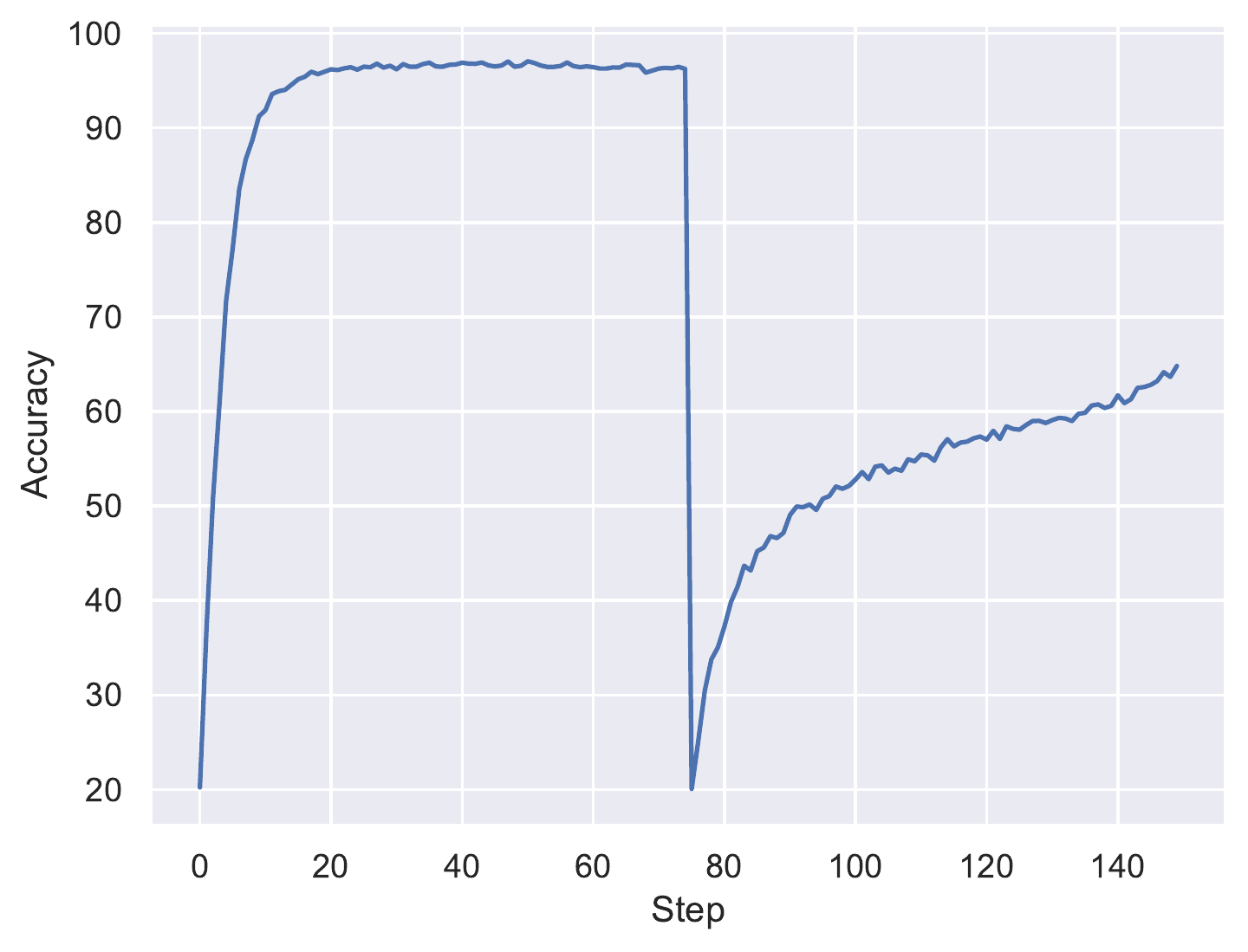}
        \caption{Average test accuracy (\%) of the SRWM-based model 
        as a function of the number of steps in the sequential multi-task adaptation setting (Sec.~\ref{sec:seq_multi}). A stream of datapoints is fed to the model in delayed-label fashion (Figure \ref{fig:delayed}). The datapoints are sampled from \textbf{Omniglot} until step 74 (where the accuracy drops),
        then from \textbf{Mini-ImageNet}.}
        \label{fig:adaptation}
    \end{center}
\end{figure}

In the end, we successfully trained both the DeltaNet baseline and the SRWM on
this sequential adaptation task.
Figure \ref{fig:adaptation} shows the evolution of the SRWM's test time accuracy as it gets more inputs.
In this test setting, the model starts with receiving a stream of samples from Omniglot.
At step 74, the task changes; the model now has to classify images sampled from Mini-ImageNet.
The accuracy obviously drops due to this change, since the model can not know which class the new datapoint belongs to, but it effectively adapts itself and starts learning the second task. Table \ref{tab:seq_adapt} compares the DeltaNet to the SRWM.
While their performance is similar in the single task scenario (Table \ref{tab:few_shot}),
the SRWM achieves a higher accuracy in this multi-task setting, demonstrating its rapid adaptation capability.
In Appendix \ref{app:train_delayed},
we also provide results for the test scenario with the reversed testing order, i.e., Mini-ImageNet then Omniglot.

\subsection{Multi-Task Reinforcement Learning (RL)}
\label{sec:rl_exp}
\begin{figure}[t]
    \begin{center}
        \includegraphics[width=.85\columnwidth]{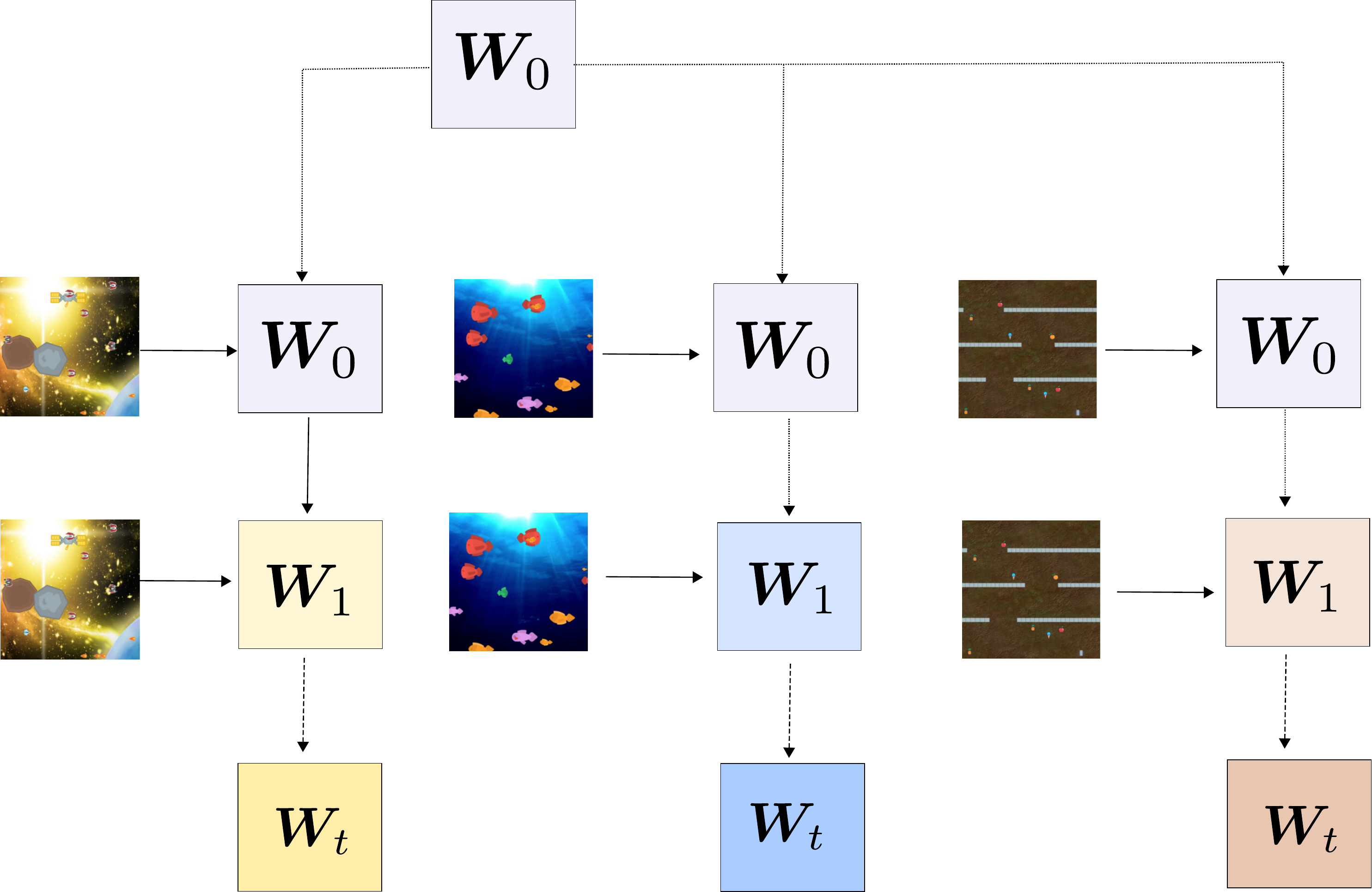}
        \caption{Illustration for multi-task RL. The initial weight matrix $\mW_0$ is common to all tasks and episodes. The effective weight matrix is a function of task/episode specific input streams.}
        \label{fig:multi-task}
    \end{center}
\end{figure}

We finally evaluate the proposed model in a multi-task RL setting using procedurally generated game environments of ProcGen \citep{CobbeHHS20}.
The corresponding setting is illustrated in Figure \ref{fig:multi-task}.
Unlike other popular game environments such as Atari \citep{BellemareOGTM16},
ProcGen provides various procedurally different levels. This allows for creating clean train/test splits.
In addition, working with diverse levels is especially
relevant to our setting, as we wish to build models which are
adaptive to changes across game types, as well as diversity within the same game.

\paragraph{General Settings.}
In our main experiment, we jointly train on 6 environments, namely, \textit{Bigfish}, \textit{Fruitbot}, \textit{Maze}, \textit{Leaper}, \textit{Plunder}, and \textit{Starpilot}
in the \textbf{easy} distribution.
We conduct distributed training using the standard IMPALA \citep{EspeholtSMSMWDF18} architecture implemented in \texttt{Torchbeast} \citep{kuttler2019torchbeast}.
We use 48 actors (i.e.~8 actors per environment).
All our models use the common \textit{large} architecture of \citet{EspeholtSMSMWDF18} which consists of a 15-layer residual convolutional vision model.
They differ from each other by 
the ``memory'' module inserted between the vision stem and the output layer.
In addition to our SRWM model, we train the baseline IMPALA feed-forward and LSTM \citep{gers2000learning, hochreiter1997long} models,
as well as two additional baselines: the DeltaNet \citep{schlag2021linear} and a ``Fake SR'' model which is the SRWM model
without the self-modification mechanism.
(i.e.~we only keep the ``y''-part in Eq.~\ref{eq:srm_start}).
We set a backpropagation span of 50 steps to train self-modification\footnote{Truncated backpropagation through time \citep{williams1990efficient} can be used to train the SRWM thanks to the additive nature of Eq.~\ref{eq:srm_end}.} as well as LSTM and DeltaNet baselines.
In all cases, the memory states (including the SRWM weight changes) are only reset at episode boundaries for all stateful models (LSTM, DeltaNet, and SRWM).
The LSTM model has one layer with 256 nodes as in the IMPALA baseline.
Both DeltaNet and SRWM have two layers with a hidden size of 128,
following the setting used by previous work \citep{irie2021going} training the DeltaNet on Atari.

These 6 environments are known for not explicitly requiring ``memory'' to perform the task (\citet{CobbeHHS20}; we also confirm this trend by comparison to our baseline feed-forward and LSTM RNN models).
In principle, this allows for evaluating the effect of self-modifications in isolation (although it is difficult to completely dissociate self-modification from the concept of ``memory'').
We jointly train on 6 environments in the easy distribution for a total of 300\,M steps (ca.~50\,M per environment).
In Appendix \ref{app:proc_mem}, we also present an extra experiment using 4 environments from the \textit{memory} distribution (\textit{Dodgeball}, \textit{Heist}, \textit{Maze}, \textit{Miner}) to evaluate our models also in partially observable settings, confirming the effectiveness of SRWM.

\paragraph{Train/Test split.}
Following \citet{CobbeHHS20},
we use 200 levels (level ID 0 to 199) to train in the easy distribution.
For evaluation, instead of randomly sampling the test levels as is commonly done for ProcGen, 
we consistently use the same set of 3 distinct test splits for all models.
Each of our test splits contains 200 levels,
respectively including levels 1000 to 1199, 1200 to 1399, and 1400 to 1599.
We use 3 test splits and report an average score to take into account the performance variability across the choice of test levels.
The performance on the training set is computed using all 200 training levels.
We train each model three times, reporting training performance
averaged over three runs, and test performance over 9 data points (3 test splits for 3 training runs).

\begin{table}[h]
\small
\setlength{\tabcolsep}{0.4em}
\caption{ProcGen normalised aggregated scores (multiplied by 100) over \textbf{6 environments} (Bigfish, Fruitbot, Maze, Leaper, Plunder, and Starpilot) in the \textbf{easy distribution}.
The models are trained in a multi-task setting.
Normalisation constants are taken from the original ProcGen paper \citep{CobbeHHS20}.
Results are computed from 3 independent training runs for 300\,M steps
in the easy distribution.
The test scores are averaged over 3 distinct sets of 200 fixed test levels (i.e.,~the mean/std computed from 9 data points).
For further details, see tables in Appendix \ref{app:tables} where we provide scores obtained for each game.
Results on memory distribution can be found in Appendix \ref{app:proc_mem}.
The number of trainable parameters are 626\,K for the
feedforward baseline (FF), 959\,K for Fake SR, 1.2\,M for LSTM, 1.05\,M for DeltaNet and 968\,K for SRWM.
}
\label{tab:norm_easy_multi}
\begin{center}
\begin{tabular}{lrrrrr}
\toprule
  & \multicolumn{1}{c}{FF} & \multicolumn{1}{c}{LSTM} & \multicolumn{1}{c}{Fake SR} & \multicolumn{1}{c}{DeltaNet} & \multicolumn{1}{c}{SRWM} \\ \cmidrule(r){2-6}
Train  & 22.5 (2.6) & 28.3 (1.4) & 27.0 (1.8) & \textbf{35.0} (1.6) &  34.6 (1.8)  \\  
Test   &  16.4 (1.6) & 15.7 (1.6) & 15.3 (1.9) & 18.6 (1.7) &   \textbf{20.0} (1.8) \\
\bottomrule
\end{tabular}
\end{center}
\end{table}

\paragraph{Overall Performance.}
Table \ref{tab:norm_easy_multi} shows the aggregated normalised scores.
First of all, comparing the LSTM and feed-forward baselines,
we confirm \citet{CobbeHHS20}'s finding that the LSTM layer does not provide
any improvements regarding the test performance (while some improvements
are obtained on the train set).
The two fast weight models (which can adapt to each task based on the task specific inputs), DeltaNet and SRWM,
clearly outperform the feedforward and LSTM baselines, as well as the Fake SR model, which is the SRWM without self-modification.
The SRWM achieves a slightly better test score
than the DeltaNet.
Overall, the proposed very generic SRWM based model achieves very competitive performance.

\paragraph{Comparison to Expert Models.}
We observe that the performance gains achieved by the SRWM
over the baselines are particularly large for two of the environments,
\textit{Bigfish} and \textit{Starpilot}.
Here we study these two cases in isolation.
In Table \ref{tab:expert}, we compare  multi-task agents presented above to expert agents trained specifically on one environment for 50\,M steps.
On \textit{Starpilot}, we observe that the self-modification mechanism yields improvements even in the single task case.
The case of \textit{Bigfish} is more interesting:
the performances of the agents with and without self-modification capability are close in the expert training case.
However, the self-modifying agent achieves a much better score in the multi-task setting, where the baseline agent's performance drops by a large margin.
This indicates the usefulness of the SRWM's ability to adapt itself to each environment in the multi-task scenario.

\begin{table}[h]
\setlength{\tabcolsep}{0.4em}
\caption{Comparison between multi-task vs expert agent performance.
Raw scores obtained in the easy distribution of ProcGen.
}
\label{tab:expert}
\begin{center}
\small
\begin{tabular}{rlrrrrr}
\toprule
 &  & \multicolumn{4}{c}{Weight Update} \\ 
 & & \multicolumn{2}{c}{No (Fake SR)} &  \multicolumn{2}{c}{Yes (SRWM)} \\
 \cmidrule(r){3-4} \cmidrule(r){5-6}
Env         & Split & Multi-6 & Expert & Multi-6 & Expert\\ \midrule
\multirow{2}{*}{Bigfish}   & Train & 11.6 (5.7) & 28.9 (0.9)  & 20.1 (2.4) &  28.5 (1.2) \\
          & Test   & 4.7 (2.4) & 15.8 (1.7)  & 9.0 (2.0) &  14.2 (2.0) \\ \midrule
\multirow{2}{*}{Starpilot} & Train & 55.0 (1.3) & 59.8 (0.7)  &  61.3 (2.0) & 64.0 (1.9)  \\
          & Test & 49.6 (2.1)  & 52.9 (1.2)  &  54.6 (2.4) &  57.3 (1.6) \\  
\bottomrule
\end{tabular}
\end{center}
\end{table}

\paragraph{Ablation on State Reset.}
The SRWM models presented above are trained by
carrying over the weight modifications across entire episodes whose lengths are variable---often episodes are getting longer during training as the agent becomes better at the task.
We were initially uncertain about the empirical stability of
the dynamics described by Eqs.~\ref{eq:srm_start}-\ref{eq:srm_end}
in such a scenario.
As an ablation study, we trained and evaluated an SRWM agent by resetting the weight update after every fixed time span (whose length was the backpropagation span).
Such models failed to leverage the SRWM mechanism,
obtaining scores of 28.5 (1.2) and 16.1 (2.2) on the train and test splits respectively, similar to those of the baseline
without self-modification (Table \ref{tab:norm_easy_multi}).

\section{Discussion}
Here we discuss interesting aspects (ignored in the previous sections) of the proposed SRWM and the experimental settings.

\paragraph{Reducing the ratio between learning complexity and number of time-varying variables.}
Our SRWM inherits a remarkable property of the 1993 system \citep{schmidhuber1993reducing}: it has many more temporal variables under massively parallel end-to-end differentiable control than what's possible in standard RNNs of the same size: $O(d^2)$ instead of $O(d)$, where $d$ is the number of hidden units.

\paragraph{Interpretability.}
In general, interpreting NNs is not straightforward.
The values of $\beta_t$ of Eq.~\ref{eq:srm_end} in the RL setting intuitively define the strength
of the weight modifications.
We observe that the values of all four components of $\beta_t$ vary between 0.50 and 0.65 depending on the input, instead of covering the full range of sigmoid values between 0 and 1.
We find it difficult to derive any further interpretation beyond these statistics.
Note, however, that $\beta_t$ by itself does not fully describe the self-modification effects of Eq.~\ref{eq:srm_end},
which also depend on the actual values of key and query.

\paragraph{Implementation/Limitation.}
Similar to recent works on fast weight programmers \citep{schlag2021linear, katharopoulos2020transformers, irie2021going},
our SRWM is implemented as a custom CUDA kernel.
While this approach yields competitive computation time and memory-efficient custom backpropagation, its flexibility is limited.
For instance, to replace all weight matrices in an RL agent's vision module or in the feature extractor for few-shot learning,
a custom implementation for convolution would be required,
although in principle the SRWM above could replace any
regular weight matrix.
In constrast, vision models that are fully based on linear layers such as the MLP-Mixer \citep{tolstikhin2021mlp} can be straightforwardly combined with the SRWM.
Regarding speed, the feedforward and LSTM baselines process about 3,500 steps per second,
while DeltaNet and SRWM do 2,300 and 1,700 steps per second respectively on a single P100 GPU in the RL experiments which require slow state copying due to separate interaction and training modes.
In supervised few-shot learning settings,
the speeds of LSTM, DeltaNet and SRWM
are comparable.
With a batch size of 128, they process about 8,000 images per second, using the same Conv-4 backend on 1-shot Omniglot on a single P100 GPU.

\paragraph{Limitation of this paper's scope.}
The main tasks in our experiments are limited to those solvable by feed-forward NNs: image classification and RL in fully observable environments.
Appendix
\ref{app:proc_mem} presents promising preliminary experiments on multi-task RL in partially observable environments using the memory distribution of ProcGen.
We found that augmenting the DeltaNet (which already has a short-term memory) with an SRWM by replacing the slow weight matrix in the DeltaNet by an SRWM yields performance improvements.
Further experiments are needed to test the SRWM on tasks which themselves require sequence processing, while
involving adaptation to different types or domains of sequences, e.g., automated domain adaptation in language modelling \citep{irie18:radmm, lazaridou2021pitfalls}.\looseness=-1

\paragraph{Other perspectives.}
While we motivated our work purely from the perspective of context-adaptive self-modifying NNs, never-ending self-reconfiguration could also be motivated by analogy to ``livewired'' synaptic connections in biological neural networks \citep{eagleman2020livewired}.

\section{Related Work}
\label{sec:rel}
\paragraph{Original Self-Referential Weight Matrix.}
The \textit{original} SRWM
was proposed in the '90s
as a framework for self-improving recurrent NNs \citep{Schmidhuber:92selfref, Schmidhuber:93selfrefann, Schmidhuber:93selfreficann, Schmidhuber:93selfrefieee}.
Such an RNN has special input and output units to directly address, read, and modify any of its own current weights through an index for each weight of its weight matrix (i.e., 
for a weight matrix with an input/output dimension $N$, the weight index ranges from 0 to $N^2-1$ which is encoded as a binary vector).
In contrast, our self-modification is based on \textit{key/value associations}, i.e., to encode a WM modification, our NN generates a key vector, value vectors, and a temporary learning rate which allows for the rapid modification of an entire rank at a time ~\citep{Schmidhuber:91fastweights,schmidhuber1993reducing}. 
This design is reinforced by the recent success of
linear Transformers and fast weight programmers \citep{katharopoulos2020transformers, schlag2021linear, irie2021going}.
In this sense, our SRWM is a \textit{modern} approach to self-modification,
even if the use of outer products to parameterise fast weight generation itself is not (e.g.~\citet{Schmidhuber:91fastweights, schmidhuber1993reducing}).

\paragraph{Other Self-Modifying Neural Networks.}
There are also more recent works on self-modifying NNs.
Neuromodulated plasticity is a Hebbian-style self-modification \citep{miconi18a, miconi2018backpropamine, Schmidgall20, NajarroR20} which also makes use of outer products to generate a modulation term which is added to the base weights.
The corresponding computations can also be interpreted as
key/value/query association operations.
However, the key, query, and value patterns are hard-coded to be one of the input/output pairs
of the corresponding layer at each time step.
While this circumvents the necessity to allocate parameters for generating those vectors,
it is known that the resulting program can be expressed in terms of an unnormalised attention~\citep{schmidhuber1993reducing}
over the past outputs \citep{ba2016using}.
In contrast, in our model, all these patterns are arbitrary as they are generated from learned 
transformations whose parameters are themselves self-modifying.

\paragraph{Hierarchical Fast Weight Programmers.}
As reviewed in Sec.~\ref{sec:background},
an FWP is an NN which learns to generate, update, and maintain weights of another NN.
However, a typical FWP has a slow NN with a weight matrix that remains fixed after training.
Previous work \citep{irie2021going} has proposed to go one step further by parameterising the slow weights
in the Delta Net with another FWP to obtain the DeltaDelta Net. 
However, such a hierarchy has no end, as the highest level programmer
would still have a fixed weight matrix.
In this work, we follow the spirit of early work \citep{Schmidhuber:92selfref, Schmidhuber:93selfrefann, Schmidhuber:93selfreficann, Schmidhuber:93selfrefieee, schmidhuber1993reducing} and
collapse these potentially hierarchical meta-levels into one single self-referential weight matrix.

\paragraph{Fixed Weight Meta-RNNs.}
Learning learning dynamics using a fixed-weight NN (typically an RNN) has become a common
approach \citep{hochreiter2001learning, cotter1990fixed, cotter1991learning, SantoroBBWL16, wang2016learning, duan2016rl, MunkhdalaiY17, MunkhdalaiSWT19, kirsch2020meta, sandler21}.
A truly self-referential weight matrix, however, would allow for modifying {\em all} of its own components. The only thing that's trained by gradient descent are the SRWM's {\em initial} weights at the beginning---all of them, however, may rapidly change during sequence processing, in a way that's driven by the SRWM itself. 

\paragraph{Dynamic Evaluation.}
Auto-regressive language modelling is
an interesting task  
in the context of sequence processing with error feedback.
In the standard testing scenario, language models receive a label feedback after each prediction at each time step.
Such feedback may not only act as error feedback and influence NN activations like in fixed weight meta-RNNs: in the common method of dynamic evaluation \citep{mikolov2010recurrent, KrauseK0R18}, it is also used to adapt the weights at test time using the fixed gradient descent algorithm.
A self-referential approach could also adapt the corresponding adaptation algorithm. 

\paragraph{Recursive Self-Improvements.}
Beyond the scope of NNs,
recursive self-modification \citep{good1965, Eden2013} is of general interest when considering
autonomous, self-improving machines \citep{Schmidhuber:87long, Schmidhuber:04agi, wang2007logic, nivel2009self, SteunebrinkTS16, wang2018self}.
While we proposed a WM that can recursively modify itself, our first set of experiments is limited to
studying its practical ability on well known supervised and reinforcement learning tasks.
In future work, we intend to define additional tasks for specifically measuring the ability to self-improve.
% by using the proposed SRWM as a backbone.
This work may also have to consider 
standard limitations of NNs
such as their lack of systematic generalisation \citep{fodor1988connectionism}, including length generalisation (e.g., see recent work on Transformers \citep{csordas2021neural}), in
the context of self-improving NNs.
For example, in the long run, there is no guarantee that the SRWM's self-modifications preserve the original objective function it is trained for \citep{hubinger2019risks}.\looseness=-1

\section{Conclusion}
We proposed a new type of self-referential weight matrix (SRWM) with a modern mechanism for self-modification.
Our self-modifying neural networks (NNs) learn to generate patterns of keys and values and learning rates, translating these patterns into rapid changes of their own weight matrix through sequential outer products and invocations of the delta update rule.
In a set of three experiments, we demonstrated that our generic SRWM is practical and performs well in both supervised few-shot learning and multi-task reinforcement learning settings, using procedurally generated game environments. Our promising results encourage further investigations of self-improving NNs.

\section*{Acknowledgements}
We would like to thank Karl Cobbe for answering some practical questions about ProcGen.
Kazuki Irie wishes to thank Anand Gopalakrishnan for letting him know about ProcGen.
This research was partially funded by ERC Advanced grant no: 742870, project AlgoRNN,
and by Swiss National Science Foundation grant no: 200021\_192356, project NEUSYM.
We are thankful for hardware donations from NVIDIA \& IBM. The resources used for the project were partially provided by Swiss National Supercomputing Centre (CSCS) project d115.

\bibliography{paper}
\bibliographystyle{icml2022}

%%%%%%%%%%%%%%%%%%%%%%%%%%%%%%%%%%%%%%%%%%%%%%%%%%%%%%%%%%%%%%%%%%%%%%%%%%%%%%%
%%%%%%%%%%%%%%%%%%%%%%%%%%%%%%%%%%%%%%%%%%%%%%%%%%%%%%%%%%%%%%%%%%%%%%%%%%%%%%%
% APPENDIX
%%%%%%%%%%%%%%%%%%%%%%%%%%%%%%%%%%%%%%%%%%%%%%%%%%%%%%%%%%%%%%%%%%%%%%%%%%%%%%%
%%%%%%%%%%%%%%%%%%%%%%%%%%%%%%%%%%%%%%%%%%%%%%%%%%%%%%%%%%%%%%%%%%%%%%%%%%%%%%%
\newpage
\appendix
\onecolumn

% You can have as much text here as you want. The main body must be at most $8$ pages long.
% For the final version, one more page can be added.
% If you want, you can use an appendix like this one, even using the one-column format.
%%%%%%%%%%%%%%%%%%%%%%%%%%%%%%%%%%%%%%%%%%%%%%%%%%%%%%%%%%%%%%%%%%%%%%%%%%%%%%%
%%%%%%%%%%%%%%%%%%%%%%%%%%%%%%%%%%%%%%%%%%%%%%%%%%%%%%%%%%%%%%%%%%%%%%%%%%%%%%%

\section{SRWM Model Details}
\label{app:model}
\paragraph{Equations for the four-learning rate case.}
In Sec.~\ref{sec:srm}, for the purpose of clarity,
we presented the equations for our SRWM
model in the case where we only have a single learning rate $\beta_t$.
Here we provide a complete description of an SRWM with a separate self-invented learning rate for each component.
As we noted in Sec.~\ref{sec:srm}, the SRWM can be split into sub-matrices: $\mW_{t-1} = [\mW^y_{t-1}, \mW^q_{t-1}, \mW^k_{t-1}, \mW^\beta_{t-1}]$ according to the sub-components used to
produce $\vy_t$, $\vq_t$, $\vk_t$, and $\beta_t$ in Eq.~\ref{eq:srm_start}.
In case where we use separate learning rates, we need
separate equations to describe the update of each sub-matrix.
For example, for the ``y''-part $\mW^y_{t-1}$, while keeping the same equation
for the first projection (Eq.~\ref{eq:srm_start}), the rest becomes:
\begin{align}
\vy^k_t &= \mW^y_{t-1} \phi(\vk_t) \\
\vy^q_t &= \mW^y_{t-1} \phi(\vq_t) \\
\mW^y_{t} &= \mW^y_{t-1} + \sigma(\beta_{y, t})(\vy^q_t - \vy^k_t) \otimes \phi(\vk_t)
\end{align}
where $\vy^k_t$ and $\vy^q_t$ are the ``y''-part of $\bar{\vv}_t$ and $\vv_t$
in Eq.~\ref{eq:srm_key} and \ref{eq:srm_query} respectively,
and $\beta_{y, t} \in \mathbb{R}$ is one of four learning rates
dedicated to the ``y''-part.
The equations for other sub-matrices $\mW^q_{t-1}$, $\mW^k_{t-1}$, $\mW^\beta_{t-1}$
are analogous.

\paragraph{Use of multiple heads.}
We inserted the SRWM between other layers with learned parameters and configurable dimensionalities.
This allows for efficient computation using multiple heads as follows.
Given a number of heads $H$ used in the SRWM layer, the model dimensions are configured such that the input dimension to an SRWM layer $d_\text{in}$ is divisible by $H$.
The input is then split into $H$ equally sized components, and each head executes separate SRWM operations (Eqs.~\ref{eq:srm_start}-\ref{eq:srm_end}) on one of the input components.
In consequence, an SRWM has fewer parameters than a DeltaNet with the same model hyper-parameters.
For example, if $d_\text{in} = d_\text{key}$, the common head dimension is $d = d_\text{in} // H$, the parameter shape of key projection in the SRWM is ($H$, $d$, $d$) while it is ($d_\text{in}$, $d_\text{in}$) = ($H * d$, $H * d$) for the DeltaNet.
This is an important remark as 
for some tasks (such as language modelling),
the number of parameters can be the dominant factor for good performance.
% while the SRWM with the same hyper-parameters as the DeltaNet results in a much smaller model.
If the input size of the SRWM layer is not configurable, this option has to be disabled and a single head version should be used.

\section{Experimental Details for Few-Shot Learning}
\label{app:fs_exp}

\subsection{Datasets}
We conduct few-shot image classification experiments using the Omniglot \citep{lake2015human},
and Mini-ImageNet \citep{VinyalsBLKW16, RaviL17} datasets.
Extra experiments using the Fewshot-CIFAR100 dataset (FC100 for short; \citet{OreshkinLL18}) 
are presented in Appendix \ref{app:extra_fsl_exp} below.
We use \texttt{torchmeta} by \citet{deleu2019torchmeta} which implements all common settings used with these datasets. 
For each dataset, classes are split into train, validation and test for few-shot learning settings.

Omniglot images are grayscale hand-written characters from 50 different alphabets, and the dataset contains 1632 different classes with 20 examples per class.
The original setting \citep{lake2015human} splits these 1632 classes into 1200 for training and 
432 for testing without validation set.
Instead, we use \citet{VinyalsBLKW16}'s 1028/172/432-split for the train/validation/test set
and their data augmentation methods based on rotation (90, 180, and 270 degrees).
The images are typically resized to $28\times 28$.
In the sequential multi-task experiments
(Sec.~\ref{sec:seq_multi}), to jointly train on Omniglot and Mini-ImageNet, we resized Omniglot images to $84\times 84$ and duplicated the channels to match the number of RGB color channels for Mini-ImageNet, such that the same feature extractor (here, Conv-4) can process both of them.

Mini-ImageNet contains 100 color image classes with 600 examples for each class.
They are typically resized to $84\times 84$.
The standard class train/valid/test splits of 64/16/20 are used \citep{RaviL17}.

FC100 is based on CIFAR100 \citep{krizhevsky}.
100 color image classes (600 images per class, each of size $32\times 32$) are split into train/valid/test classes of 60/20/20 \citep{OreshkinLL18}.

\subsection{Model and Training Details}
\paragraph{Vision feature extractors.}
We mainly evaluated our models on few-shot learning using the standard convolution-based vision feature extractor, Conv-4, proposed by \citet{VinyalsBLKW16}.
It has four blocks, each consisting of one $3 \times 3$ 2D-convolutional layer, batch normalisation, max-pooling of size 2, and a ReLU activation layer.
We use 32 channels in each layer.
The feature dimension of this encoder's output is 64, 800, and 128 for Omniglot, Mini-ImageNet, and FC100, respectively.
Dropout \citep{hanson1990stochastic, srivastava2014dropout} is applied after each max-pooling layer.

\paragraph{Sequence processing components.}
Both the SRWM and DeltaNet-based models used in this work follow the basic Transformer architecture \citep{trafo} where the self-attention layers are replaced by the corresponding DeltaNet (Eqs.~\ref{eq:proj}-\ref{eq:fw_get}) and SRWM (Eqs.~\ref{eq:srm_start}-\ref{eq:srm_end}) operations.
For supervised tasks, no activation function  is applied to $\vx_t$ in Eq.~\ref{eq:srm_start} of the SRWM, while softmax is applied for the RL experiments.

\paragraph{Training hyper-parameters.}
For Omniglot, we use two layers of size 256 using 16 computational heads and 1024 (4 * 256) dimensional feed-forward inner dimensions. We train with a learning rate of 1e-3 with a batch size of 128 for 300\,K steps and validate every 1000 steps.
For Mini-ImageNet,
we conduct hyper-parameter search for the SRWM and the DeltaNet as follows: a number of layers $l \in \{2, 3, 4\}$, a hidden size $d_\text{model} \in \{128, 256\}$,
two dropout rates $p_\text{vision}, p \in \{0.0, 0.1, 0.2, 0.3\}$ (separately for the vision and sequence processing components) and a learning rate $\eta \in \{1e-3, 3e-4, 1e-4\}$ or the standard Transformer ``warmup'' learning rate scheduling \citep{trafo}.
The number of heads is fixed to 16.
We set a feed-forward inner dimension to $d_\text{ff} = m * d_\text{model}$ where $m \in \{4, 8\}$.
For the 1-shot Mini-ImageNet, the best SRWM was obtained for $(l=3, d_\text{model}=256, d_\text{ff}=1024, p_\text{vision}=0.1, p=0.1, \eta=\text{``warmup''})$ and the best DeltaNet was obtained for $(l=4, d_\text{model}=256, d_\text{ff}=2048, p_\text{vision}=0.3, p=0.1, \eta=1e-4)$.
For the 5-shot Mini-ImageNet, the best SRWM was obtained for $(l=3, d_\text{model}=256, d_\text{ff}=2048, p_\text{vision}=0.1, p=0.1, \eta=\text{``warmup''})$ and the best DeltaNet was obtained for $(l=4, d_\text{model}=256, d_\text{ff}=2048, p_\text{vision}=0.2, p=0.0, \eta=3e-4)$.
In both cases, the number of trainable parameters is 3.6\,M for the SRWM and 5.5\,M for the DeltaNet (see comments in Appendix \ref{app:model} above).
For Mini-ImageNet, we use a batch size of 16
and train for 600\,K steps (470\,K steps roughly correspond to one epoch, i.e., covering all class combinations for the choice of 5 classes for few-shot learning).
All models are trained using the Adam optimiser.

\subsection{Evaluation Procedure}
Following the standard evaluation setting of \citet{RaviL17}, we report the mean and 95\% confidence interval computed over multiple sets of test episodes.
We use 5 different sets consisting of 16\,K random test episodes each.
However, we also note that this metric does not show the performance
variance across seeds (which is typically high on these tasks---this is not specific to our approach).
In fact, for many configurations, with some seeds, the model performance can remain around the random guessing accuracy of 20\% for the entire duration of training.

\subsection{Extra Results}
\label{app:extra_fsl_exp}
\paragraph{Few-Shot CIFAR100 (FC100).}
Here we provide extra experimental results on the FC100 dataset.
We conduct a hyper-parameter search similar to the one described above for Mini-ImageNet.
The best SRWM is obtained for $(l=3, d_\text{model}=256, d_\text{ff}=1024, p_\text{vision}=0.1, p=0.1, \eta=\text{``warmup''})$ and the best DeltaNet is obtained for $(l=3, d_\text{model}=256, d_\text{ff}=2048, p_\text{vision}=0.2, p=0.0, \eta=\text{``warmup''})$.
Table \ref{tab:fc100} shows the results.
First of all, we confirm again that the SRWM and the DeltaNet achieve similar performance.
However, our generic models underperform the TADAM \citep{OreshkinLL18} method by a large margin.\footnote{In a previous preprint of this paper, we reported 57.1\% (instead of 44.5 \%) for the Conv-4-32 backend in this setting,
due to a bug in the meta-test data shuffling process.
Details of this correction can be found in our public code on GitHub.
We re-ran all important experiments after this fix to produce the numbers presented in the present version.}
To match the vision model architecture,
we also test the SRWM using the Res-12 architecture of \citet{chen2021meta} which has four residual blocks consisting of three $3 \times 3$ convolutional layers.
The Res-12-256 architecture uses the following numbers of channels: 64, 96, 128, 256 in the respective blocks \citep{mishra2018a}, and spatial dropout \citep{TompsonGJLB15} is used.
While we achieve slight improvements through this Res-12-256 vision model, no further improvement is obtained on this task by increasing the number of channels as in the Res-12-512 architecture of TADAM.

We observe that the SRWM outperforms the ProtoNet variant (ProtoNet Cosine; \citet{OreshkinLL18}) using the cosine similarity without a trainable scaling factor inside the softmax (like in the output layer of our models), but underperforms the one with scaling (ProtoNet Cosine Scaled).
This result seems to indicate that
additional tuning specific to few-shot learning is necessary to achieve competitive performance on this more challenging task.

\begin{table}[h]
\caption{5-way 5-shot image classification experiments on \textbf{FC100}.
All numbers marked by * are taken from \citet{OreshkinLL18}.
}
\label{tab:fc100}
\begin{center}
\begin{tabular}{llr}
\toprule
Model & Backend & \multicolumn{1}{c}{Accuracy}  \\ \midrule 
ProtoNet Cosine* & Res12-512 & 40.9 $\pm$ 0.6 \\
ProtoNet Cosine Scaled* & & 51.0 $\pm$ 0.6 \\ 
TADAM*   &  &  \textbf{56.1} $\pm$ 0.4 \\  \midrule \midrule
DeltaNet   & Conv-4-32  & 43.2 $\pm$ 0.1 \\
SRWM   & Conv-4-32  & 44.5 $\pm$ 0.1 \\
       & Res-12-256 & 46.3 $\pm$ 0.1 \\
\bottomrule
\end{tabular}
\end{center}
\end{table}

\paragraph{Tuning the LSTM baseline.}
It is typically found difficult to achieve competitive performance on few-shot image classification tasks using an LSTM \citep{mishra2018a}.
Indeed, we find that for many configurations, the LSTM's performance remains at chance level.
For some seeds and with longer training, however, the LSTM can achieve competitive performance.
Table \ref{tab:few_shot_lstm} shows the best obtained results.
Lack of training stability prevents it from being a reliable baseline (e.g., we did not manage to successfully train any LSTM model on FC100), but we show that we can obtain good LSTM results in case of fully successful training.

The best LSTM configuration for Mini-ImageNet has one layer with a hidden size of 1024 for both 1-shot and 5-shot cases (7.5\,M parameters), with a vision dropout rate of 0.1.
The warmup learning rate scheduling also works well for the LSTM.
The model is trained for 1.2\,M steps in the 5-shot case (while 600\,K steps are enough for the DeltaNet and SRWM).
We use the Adam optimiser \citep{kingma2014adam} and a batch size of 16.
For Omniglot, the model has two layers with a hidden size of 512, which is trained with a learning rate of 1e-3 and a batch size of 128 for 300 K steps similarly to other models.

\begin{table*}[h]
\caption{Test accuracies (\%) of LSTM on single task, 5-way, few-shot classification on Omniglot and Mini-ImageNet.
We report 95\% confidence intervals computed over five sets of test episodes.
``Steps'' refer to training steps; each is a batch of 16 episodes.
}
\label{tab:few_shot_lstm}
\vskip 0.15in
\begin{center}
\begin{tabular}{lclcccc}
\toprule
       &   Omniglot  & \multicolumn{5}{c}{Mini-ImageNet} \\ \cmidrule(r){2-2} \cmidrule(r){3-7} 
        & \multirow{2}{*}{1-shot} & \multirow{ 2}{*}{Backend} & \multicolumn{2}{c}{1-shot} & \multicolumn{2}{c}{5-shot}  \\ \cmidrule(r){4-5} \cmidrule(r){6-7}
   &  &  & 600\,K steps & 1.2\,M steps  & 600\,K steps & 1.2\,M steps \\ \midrule
LSTM   &   96.3  $\pm$ 0.1  & Conv-4-32 & $<$ 38 & 46.6 $\pm$ 0.1 & $<$ 55 & 59.0 $\pm$ 0.1 \\
\bottomrule
\end{tabular}
\end{center}
\vskip -0.1in
\end{table*}

\paragraph{Extra results on the sequential multi-task setting.}

\begin{figure}[h]
\centering
\begin{minipage}{.432\textwidth}
  \centering
  \includegraphics[width=\columnwidth]{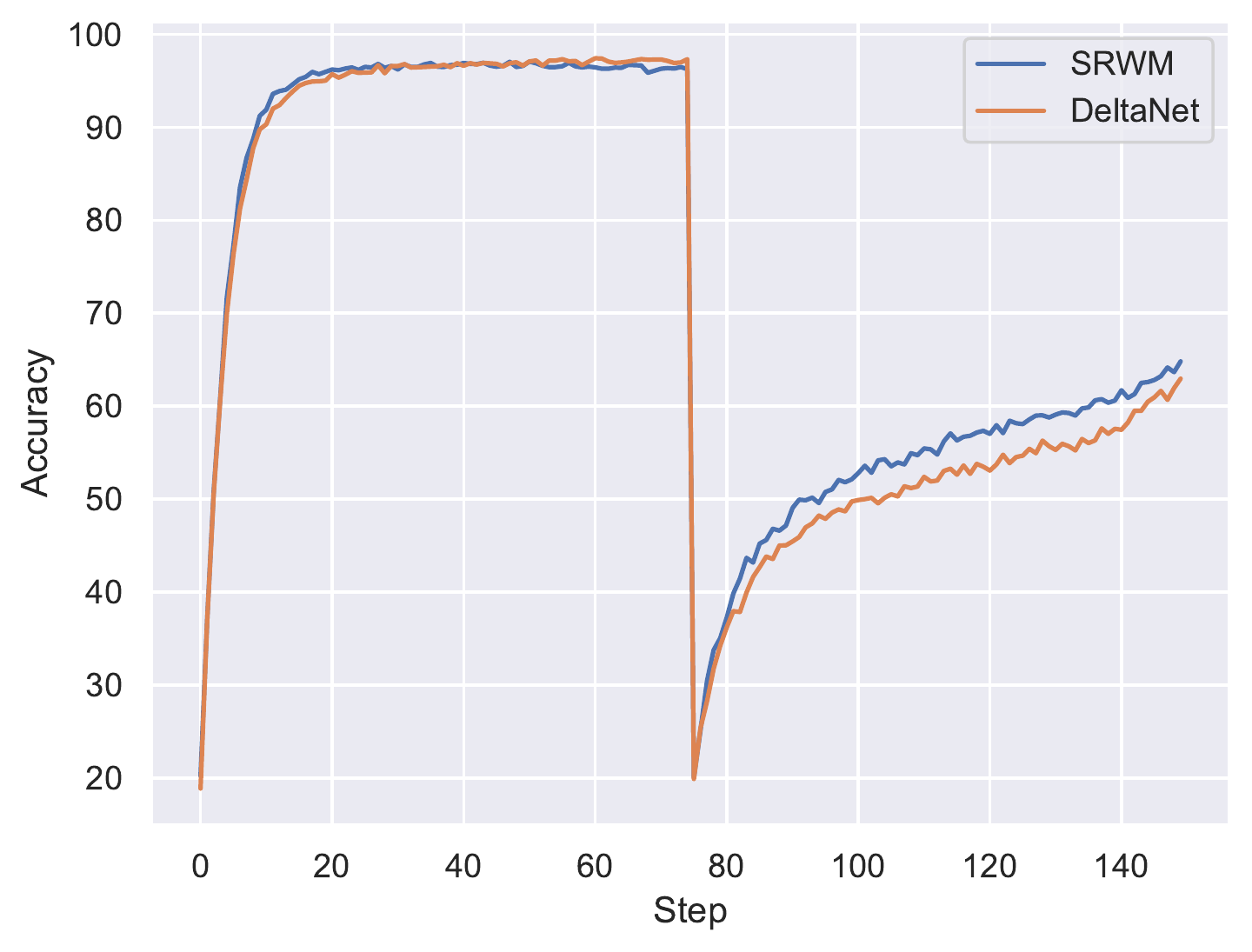}
   \vspace{-5mm}
  \caption{Average test accuracy (\%) as a function of the number of steps in the sequential multi-task adaptation setting. The datapoints are sampled from \textbf{Omniglot} until step 74 (where the accuracy drops),
then from \textbf{Mini-ImageNet}.}
  \label{fig:curves_om_first}
\end{minipage}
\hspace{5mm}
\begin{minipage}{.45\textwidth}
  \centering
  \includegraphics[width=0.98\columnwidth]{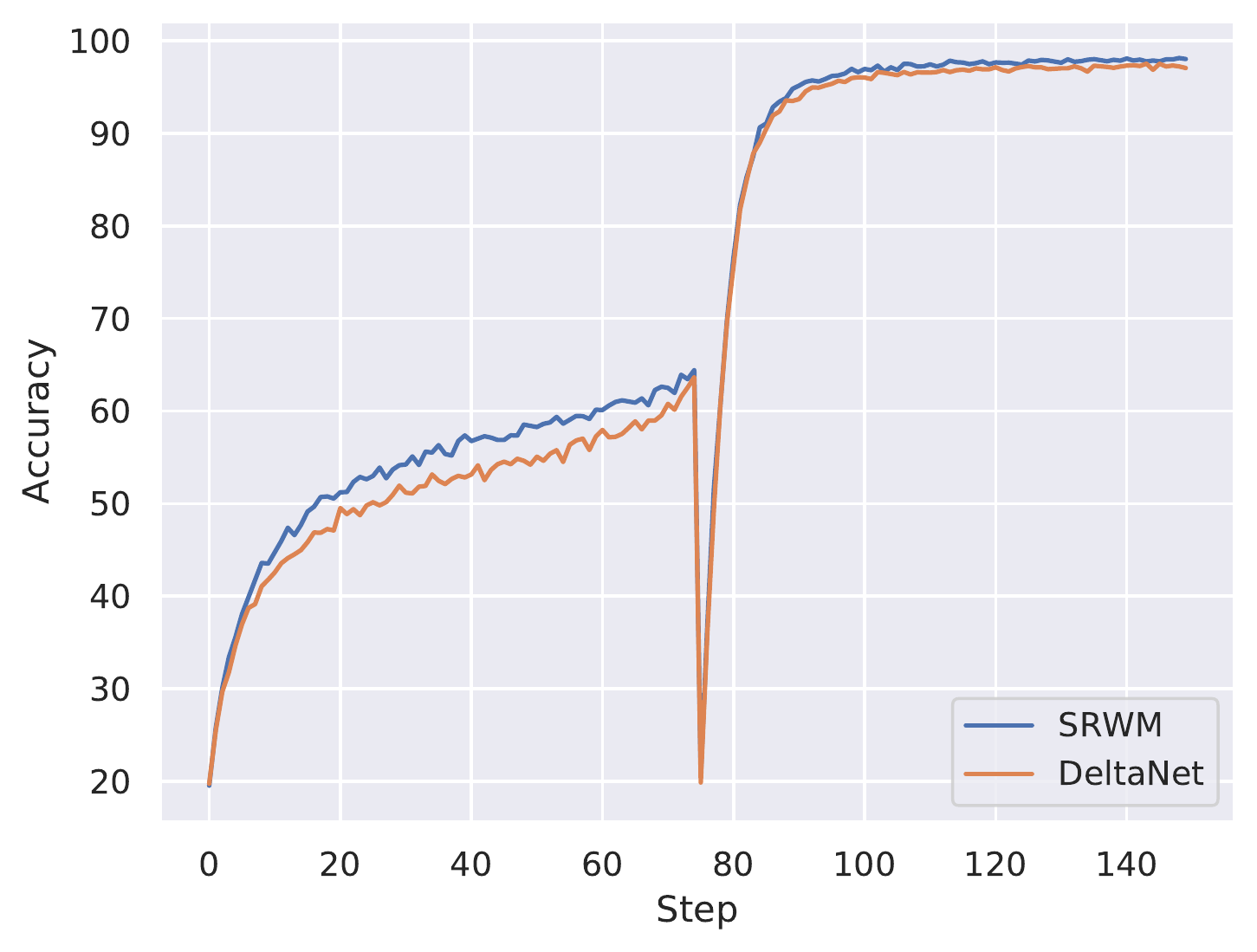}
  \vspace{-1mm}
   \caption{Average test accuracy (\%) as a function of the number of steps in the sequential multi-task adaptation setting. The datapoints are sampled from \textbf{Mini-ImageNet} until step 74 (where the accuracy drops), then from \textbf{Omniglot}.}
  \label{fig:curves_im_first}
\end{minipage}%
\end{figure}

In Table \ref{tab:seq_adapt}, we present results for sequential multi-task few-shot learning experiments where the test sequences start with Omniglot and then change to Mini-ImageNet.
Here we reverse the order of testing: Mini-ImageNet, then Omniglot.
Table \ref{tab:seq_adapt_im_first} shows the results.
We observe that improvements by the SRWM on the (second) Omniglot part are slightly larger in this ordering.
We also show the corresponding test accuracy curves in
Figures \ref{fig:curves_om_first} (Omniglot first) and \ref{fig:curves_im_first} (Mini-ImageNet first).
Note that the SRWM part of Figure \ref{fig:curves_om_first}
is already present in Figure \ref{fig:adaptation} (where the DeltaNet part is omitted for clarity).

\begin{table*}[h]
\caption{Total and instance-level accuracies (\%) for \textbf{sequential multi-task few-shot learning}
experiments (Sec.~\ref{sec:seq_multi}).
Regarding the instance-level accuracy, column $k \in \{1, 2, 3, 5, 10\}$ shows the percentage of correctly predicted $k$-th instances from each class.
In this test time scenario the model is first tasked to learn to predict \textbf{Mini-ImageNet, then Omniglot}.
}
\label{tab:seq_adapt_im_first}
% \vskip 0.15in
\begin{center}
\begin{tabular}{rr|rrrrr|r}
\toprule
Task & Model & 1 & 2 & 3 & 5 & 10 & Total \\ \midrule
\multirow{2}{*}{Mini-ImageNet}  & DeltaNet & 20.3 & 44.7 & 48.9 & 51.8 & 54.8 & 50.7 \\
         & SRWM & 22.4 & 46.4 & 50.3 & 54.3 & 57.9 & \textbf{53.5} \\  
     \midrule    
\multirow{2}{*}{Omniglot}  & DeltaNet & 39.2 & 90.9 & 93.8 & 95.7 & 96.9 & 92.2 \\
         & SRWM & 39.6 & 91.2 & 94.6 & 96.6 & 97.6 & \textbf{92.9} \\
\bottomrule
\end{tabular}
\end{center}
\end{table*}

\subsection{Training in Delayed Label Setting}
\label{app:train_delayed}
\begin{figure}[h]
\centering
\begin{minipage}{.42\textwidth}
  \centering
  \includegraphics[width=\columnwidth]{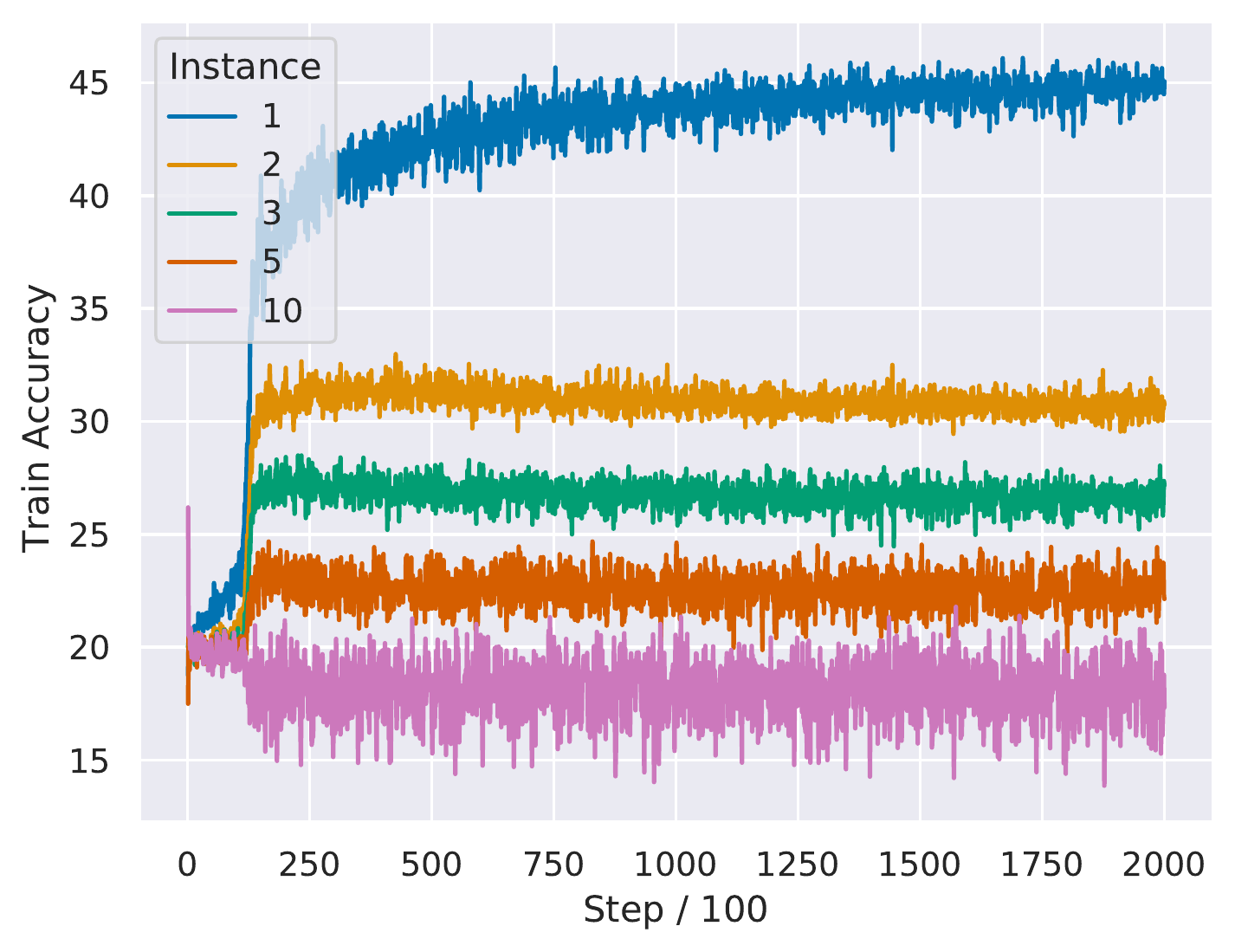}
   \vspace{-4mm}
  \caption{Example \textbf{unsuccessful} training on the sequential multi-task few-shot learning task. Average training accuracies for different instances are shown for the Omniglot part.}
  \label{fig:unsuccessful}
\end{minipage}
\hspace{5mm}
\begin{minipage}{.45\textwidth}
  \centering
   \vspace{-5mm}
  \includegraphics[width=0.98\columnwidth]{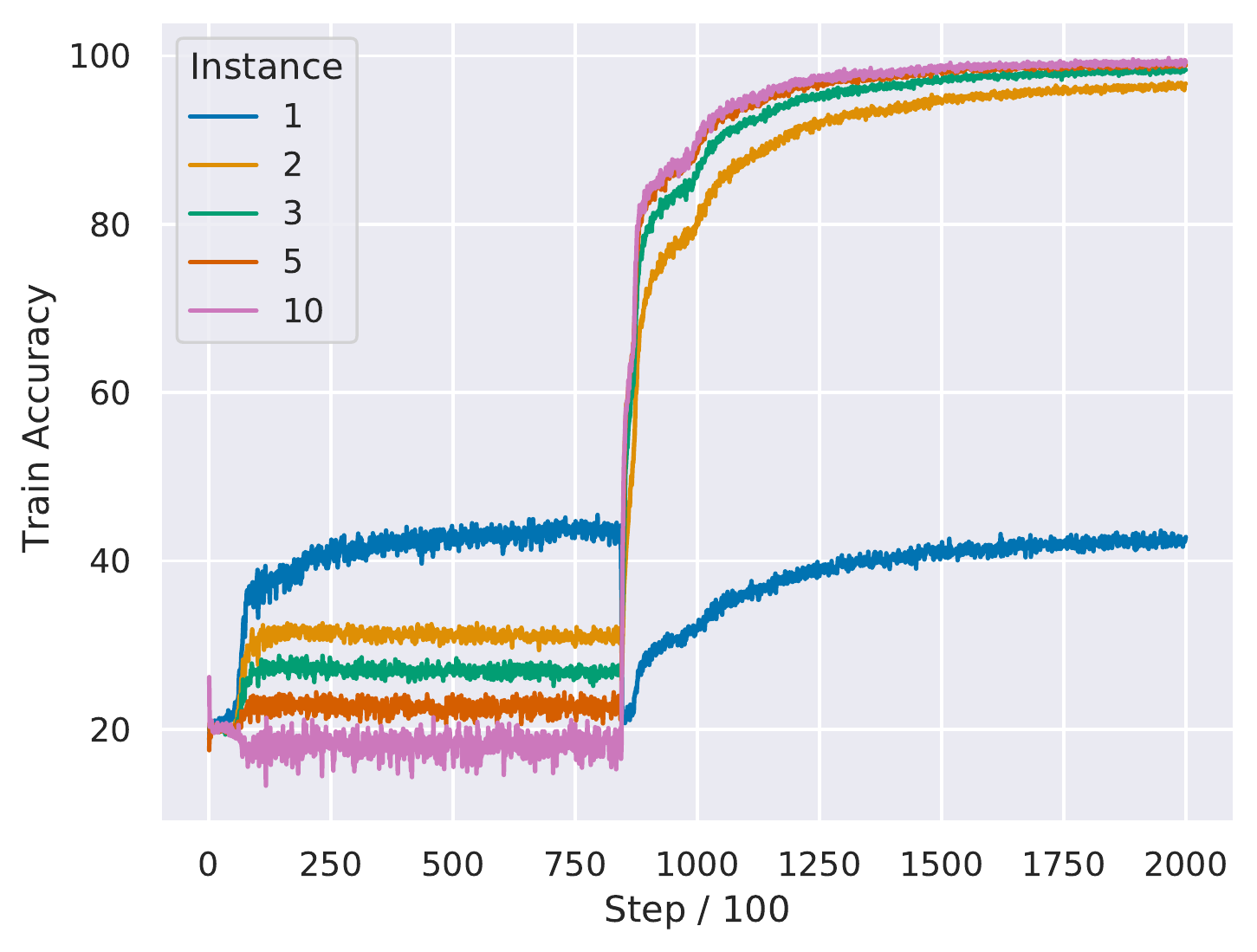}
  % \vspace{-3mm}
   \caption{Example \textbf{successful} training on the sequential multi-task few-shot learning task. Average training accuracies for different instances are shown for the Omniglot part.}
  \label{fig:successful}
\end{minipage}%
\end{figure}

In the sequential multi-task setting
of Sec.~\ref{sec:seq_multi},
for both the SRWM and the DeltaNet, we set the number of layers to three, with a hidden layer size of 256 using 16 computational heads, and a feed-forward dimension of 256.
We train with a batch size of 32 (crucial for successful training) and a learning rate of 3e-4.
During training, each Omniglot and Mini-ImageNet segment is constructed using up to 15 examples per class (which yields the maximum segment length of 75 images in this 5-way classification setting).
For each batch, the number of positions to be trimmed is randomly sampled between 1 and 60 for Omniglot and Mini-ImageNet segments during training, to prevent training sequences from having always the same number of examples per class.
As described in the main text (Sec.~\ref{sec:seq_multi}),
bad configurations or unlucky seeds get stuck with sub-optimal behaviour.
Here we show some training curves of such  behaviours in Figure \ref{fig:unsuccessful}, as well as successful ones in Figure \ref{fig:successful}.

%   --batch_size 32 \
%   --grad_cummulate 2 \
%   --num_layer 3 \
%   --hidden_size 256 \
%   --n_head 16 \
%   --k_shot 15 \
%   --max_trim 60 \
%   --ff_factor 1 \
%   --learning_rate 3e-4 \

\section{Additional Results for Reinforcement Learning Experiments}

\subsection{Experiments on ProcGen Memory Distribution}
\label{app:proc_mem}

Sec.~\ref{sec:rl_exp} presents our experimental results on 6 environments in the easy distribution.
Here we present an extra experiment using 4 environments in the \textit{memory} distribution (\textit{Dodgeball}, \textit{Heist}, \textit{Maze}, \textit{Miner}) to evaluate our models also in partially observable settings.
In ProcGen memory distributions, the world size  is increased
and the observations are restricted
to a small patch of space around the agent \citep{CobbeHHS20}.
In such partially observable environments, the baseline models have to be sequence processing NNs with memory, such as the DeltaNet.
Here we are interested in augmenting such a model with an extra self-referential mechanism.
We thus replace the slow weight matrix of DeltaNet (Eq.~\ref{eq:proj}) by an SRWM (Eqs.~\ref{eq:srm_start}-\ref{eq:srm_end}).
The resulting model thus learns to generate and update a fast weight matrix as a short-term memory, while it also learns to modify itself.
We denote this model ``SR-DeltaNet.''

We train with a backpropagation span of 100 steps for
this memory distribution setting.
While we train for a total of 300\,M steps (ca.~50\,M per environment) for the joint training on 6 environments in the easy distribution,
here we train for 800\,M steps (200\,M steps per environment) on 4 environments.

Since there is no standard convention \citep{CobbeHHS20} for the memory distribution setting, we opt for training on 500 training levels (level ID 0 to 499) as recommended for the ``hard'' distribution.
For testing, we use the exact same setting as in our easy distribution setting described in Sec.~\ref{sec:rl_exp} (i.e.~3 training runs and 3 different test sets).

Table \ref{tab:norm_mem_multi} shows the results.
While having a similar parameter count, the SRWM variant
achieves a better training score than the DeltaNet baseline, while the test scores are rather close.

We note, however, that this is only a preliminary result in the memory setting, as the model size is rather small: we use the same 2-layer architecture used in the experiments with the easy distribution.
Further scaling up the model size (e.g., more layers) should lead to further performance improvements
as it would allow for handling longer contexts.

\begin{table}[t]
% \RawFloats
\caption{ProcGen normalised aggregated scores (multiplied by 100) over \textbf{4 environments} (Dodgeball, Heist, Maze, Miner) in the \textbf{memory} distribution.
The models are trained in a multi-task setting.
We use the normalisation constants for the hard distribution provided in the original ProcGen paper.
Results are derived from 3 independent training runs with 800\,M steps each.
The test scores are averaged over 3 distinct sets of 200 fixed test levels (i.e.~the mean/std computed from 9 data points).
For further details, see tables in Appendix \ref{app:tables} where we provide scores obtained for each game.
}
\label{tab:norm_mem_multi}
% \vskip 0.15in
\begin{center}
\begin{tabular}{rrr}
\toprule
  & \multicolumn{1}{c}{DeltaNet} & \multicolumn{1}{c}{SR-DeltaNet}  \\ \midrule 
Train  & 51.8 (2.6) & \textbf{59.0} (2.1) \\  
Test   & 38.0 (4.1) & \textbf{38.5} (3.2) \\
\bottomrule
\end{tabular}
\end{center}
\end{table}

\subsection{Extra result tables}
\label{app:tables}
% =================== Easy distribution =====================

\begin{table}[h]
\caption{Performance on ProcGen game environments. Multi-task training in 6 environments in the \textbf{easy distribution}.
The three SRWM variants are as follows:
\textit{True}: the SRWM model, \textit{Fake}: the SRWM without
self-modification mechanism, and \textit{Reset}: the SRWM trained
and evaluated with weight update reset.
}
\label{app:tab:multi6}
% \vskip 0.15in
\begin{center}
\begin{tabular}{rlrrrrrr}
\toprule
\multirow{2}{*}{Env} & \multirow{2}{*}{Split} & \multicolumn{1}{c}{\multirow{2}{*}{FF}}  & \multicolumn{1}{c}{\multirow{2}{*}{LSTM}} & \multicolumn{1}{c}{\multirow{2}{*}{Delta}} & \multicolumn{3}{c}{SRWM} \\ \cmidrule(r){6-8} 
 &  &   &  &  & \multicolumn{1}{c}{True} & \multicolumn{1}{c}{Fake} & \multicolumn{1}{c}{Reset} \\ \midrule
Bigfish   & Train & 8.3 (3.9) & 6.5 (2.0) & 19.6 (4.0) & \textbf{20.1} (2.4) & 11.6 (5.7) & 15.7 (2.8) \\
          & Test  &  4.3 (2.3) & 3.2 (1.1) & 7.8 (1.5) & \textbf{9.0} (2.0) & 4.7 (2.4)  & 5.8 (1.3) \\ \midrule
Fruitbot  & Train & \textbf{29.2} (0.2) & 27.8 (0.5) & 28.8 (0.9) & 28.7 (0.2) & 27.8 (1.3) & 29.2 (0.2) \\
          & Test  & \textbf{25.6} (1.1) & 24.8 (0.7) & 24.5 (1.5) & 25.5 (1.0) & 24.6 (1.2) & 25.2 (1.4) \\ \midrule
Leaper    & Train & 3.3 (0.2) & 3.3 (0.2) & \textbf{3.5} (0.4) & \textbf{3.5} (0.2) & 3.3 (0.3) & 3.4 (0.2) \\
          & Test  & 3.4 (0.4) &  \textbf{3.6} (0.4) &  3.3 (0.4) & 3.4 (0.4) & \textbf{3.6} (0.4) & 3.5 (0.3) \\ \midrule
Maze  & Train & 1.9 (0.3) & 3.1 (0.7) &  \textbf{3.8} (0.2) & 3.6 (0.5) & 3.2 (0.2) & 2.9 (0.2) \\
      & Test  & 1.4 (0.3) & 1.6 (0.4) & 1.7 (0.2) &  \textbf{1.8} (0.5) & 1.3 (0.3) & 1.5 (0.3) \\ \midrule
Plunder   & Train & 3.2 (0.2) & 3.2 (0.4) &  \textbf{3.3} (0.2) & 3.1 (0.0) & 3.1 (0.4) & 3.1 (0.1) \\
          & Test  & 3.2 (0.3)  & 2.9 (0.4) &  \textbf{3.3} (0.2) & 3.0 (0.2) & 3.1 (0.5) & 3.0 (0.2) \\ \midrule
Starpilot & Train & 57.6 (0.9) & 56.0 (1.5) & 60.3 (0.4) &  \textbf{61.3} (2.0) & 55.0 (1.3) & 55.0 (1.9) \\
          & Test  & 53.0 (1.7) & 48.3 (2.0) & 53.9 (2.4) &  \textbf{54.6} (2.4) & 49.6 (2.1) & 48.6 (1.9) \\  \midrule \midrule
Aggregated & Train & 22.5 (2.6) & 28.3 (1.4) & \textbf{35.0} (1.6) & 27.0 (1.8) & 34.6 (1.8) & 28.5 (1.2) \\
& Test & 16.4 (1.6) & 15.7 (1.6) & 18.6 (1.7) & \textbf{20.0} (1.8) & 15.3 (1.9) & 16.1 (2.2) \\
\bottomrule
\end{tabular}
\end{center}
\end{table}

% ======================= POMDP =====================
\begin{table}[h]
\caption{Performance on ProcGen game environments. Multi-task training in 4 environments in the \textbf{memory distribution}.
}
\label{app:tab:multi4_po}
% \vskip 0.15in
\begin{center}
\begin{tabular}{rlrr}
\toprule
Env & Split & DeltaNet & SRM-Delta \\ \midrule
Dodgeball & Train  &  \textbf{7.1} (0.2) &  \textbf{7.1} (0.6) \\
       & Test  &  \textbf{6.4} (0.3) & 6.2 (0.6) \\ \midrule
Heist & Train  & 1.0 (0.3) &  \textbf{1.5} (0.1)  \\
      & Test   & 0.8 (0.2) &  \textbf{1.1} (0.3)\\ \midrule
Maze & Train  & 5.3 (0.4)&  \textbf{5.9} (0.2) \\
     & Test   &  \textbf{3.3} (0.6) &  \textbf{3.3} (0.4) \\ \midrule
Miner  & Train & 32.3 (0.4) & \textbf{34.5} (0.8) \\
       & Test  & 29.2 (1.1) &  \textbf{29.4} (0.7) \\ \midrule \midrule
Aggregated  & Train & 51.8 (2.6) & 59.0(2.1) \\
                  & Test  & 38.0 (4.1) & 38.5(3.2) \\
\bottomrule
\end{tabular}
\end{center}
\end{table}

\end{document}